\pdfoutput=1

\documentclass[11pt]{article}

\usepackage{acl}

\usepackage{times}
\usepackage{latexsym}
\usepackage{enumitem}

\usepackage[T1]{fontenc}

\usepackage[utf8]{inputenc}
\usepackage{multirow}
\usepackage{makecell}

\usepackage{microtype}

\usepackage{inconsolata}

\usepackage{graphicx}
\usepackage{amsmath,amsfonts,amssymb}

\usepackage{algorithmic}
\usepackage{algorithm}

\usepackage{helvet}
\usepackage{booktabs}
\usepackage{multirow}
\usepackage{booktabs} 

\usepackage[most]{tcolorbox}
\tcbuselibrary{listings}

\title{GASim: A Graph-Accelerated Hybrid Framework for Social Simulation}

\def\correspondingauthor{\thanks{Corresponding author}}

\author{
 \textbf{Xuan Zhou\textsuperscript{1}},
 \textbf{Yanhui Sun\textsuperscript{1}},
 \textbf{Hantao Yao\textsuperscript{1}},
 \textbf{Allen He\textsuperscript{2}},  \\
 \textbf{Yongdong Zhang\textsuperscript{1}},
 \textbf{Wu Liu\textsuperscript{1}\correspondingauthor{}} 
 \\ [1ex]
 \textsuperscript{1}University of Science and Technology of China, \\
 \textsuperscript{2}BASIS International School Park Lane Harbour 
 \\ [1ex]
  \texttt{\{xuanzhou0201, yanhuis1999\}@mail.ustc.edu.cn}, \, \texttt{allenhethis@outlook.com},\\
  \texttt{\{yaohantao, zhyd73, liuwu\}@ustc.edu.cn}\\
}

\begin{document}
\maketitle
\begin{abstract}
Large-scale social simulators are essential for studying complex social patterns. Prior work explores hybrid methods to scale up simulations, combining large language models (LLM)-based agents with numerical agent-based models (ABM). However, this incurs high latency due to expensive memory retrieval and sequential ABM execution. 
To address this challenge, we propose GASim, a graph-accelerated hybrid multi-agent framework for large-scale social simulations. For core agents driven by LLM,
GASim introduces Graph-Optimized Memory (GOM) to replace intensive LLM-based retrieval pipelines with lightweight propagation over a sparse memory graph. For the majority of ordinary agents, GASim employs Graph Message Passing (GMP), substituting sequential ABM execution with parallel updates by fine-grained feature aggregation and Graph Attention Network. We further introduce Entropy-Driven Grouping (EDG) that coordinates this hybrid partitioning, leveraging information entropy to dynamically identify emergent core agents situated in information-diverse neighborhoods.
Extensive experiments show that GASim not only delivers a substantial 9.94× end-to-end speedup over the traditional hybrid framework but also consumes less than 20\% of baseline tokens, significantly reducing costs while preserving strong alignment with real-world public opinion trends. Our code is available at \url{https://github.com/Jasmine0201/GASim}.
\end{abstract}

\section{Introduction}
Large Language Models (LLMs) have demonstrated remarkable human-like performance in perception and reasoning \cite{AtomThink, HOIGen-1M, A-mem, SingleFrame}. Consequently, the LLM-based multi-agent systems provide a powerful paradigm for simulating complex social dynamics \cite{stanfordtown}, where large-scale agent populations are crucial for high-fidelity simulations and have attracted growing research interest \cite{LMAgent, largescale}.

Scaling social simulations to thousands or millions of agents often requires great computation cost of heavy distributed LLM workloads \cite{oasis, kairos}, which constrains further scalability. To mitigate this cost, hybrid frameworks, exemplified by HiSim \cite{hisim}, adopt few LLM-driven core agents to model opinion leaders, while simulating the remaining ordinary agents with numerical agent-based models (ABM) \cite{Lorenz, HK, RA}. In this setting, core agents handle perception, memory retrieval, and opinion generation via LLMs, whereas ordinary agents update opinion values with rule-based ABMs \cite{DynamiX}.

\begin{figure}[t]
    \centering
    \includegraphics[width=\columnwidth]{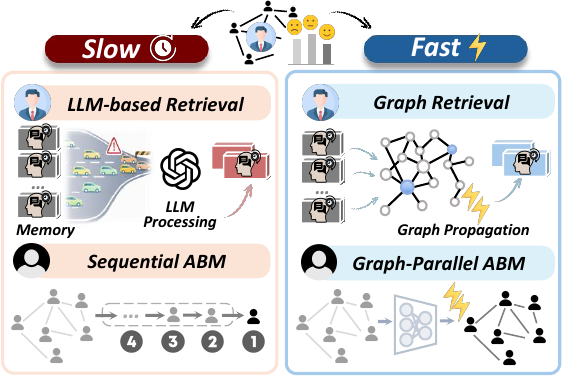}
    \caption{Comparison between the traditional hybrid framework for social simulation and our approach. Our method accelerates simulation with lightweight graph memory retrieval for core agents and parallel Agent-Based Model (ABM) execution for ordinary agents.}
    \label{fig:intro_fig}
\end{figure}

However, such hybrid frameworks fail to meet the low-latency demands of large-scale social simulation due to two key bottlenecks, as illustrated in Figure~\ref{fig:intro_fig}. First, core agents incur substantial memory retrieval overhead, as many designs rely on LLM-in-the-loop processing over large volumes of memory items \cite{agentverse, A-mem, mem0} during simulations, which dominates runtime at scale. 
Lightweight memory methods such as vector similarity search \cite{FAISS} have been explored, but they often collapse into unstructured memory as storage grows. In contrast, graph-based approaches \cite{A-mem, mem0} preserve structure but incur costly LLM-driven edge construction overhead.
Second, ordinary agents are constrained by the sequential execution of ABMs, causing runtime to scale linearly. For instance, updating one million agents would take over 100 hours for a 30-step rollout based on the experiment.

To address these challenges, we propose \textbf{GASim}, a graph-accelerated hybrid multi-agent framework for large-scale social simulations. a) For core agents, to avoid costly LLM-based processing in memory retrieval, we introduce \textbf{Graph-Optimized Memory (GOM)} that bridges lightweight similarity matching and structured memory. GOM first constructs a sparse memory graph for each agent based on similarity over perceived content, keywords, and opinion values. To enable fast and accurate retrieval, GOM casts retrieval as an optimization problem and solves it with a lightweight graph propagation algorithm, eliminating network and decoding overhead. b) For ordinary agents, to eliminate the sequential bottleneck of ABM, we design \textbf{Graph Message Passing (GMP)} to parallel agents' opinion updates. GMP computes agents’ dynamic stance features and static profile features in batched tensor operations, and leverages a Graph Attention Network \cite{gat} to update agents' opinion scores in a single forward pass. c) Moreover, noting that the traditional degree-based agent grouping fails to capture dynamically emerging opinion leaders, we propose \textbf{Entropy-Driven Grouping (EDG)}, which adaptively distinguishes opinion-leading core agents from ordinary agents based on neighborhood opinion diversity, capturing the temporal evolution of social influence.

Extensive experiments demonstrate that GASim significantly outperforms the traditional hybrid framework in efficiency. It achieves a \textbf{9.94× end-to-end speedup}, with \textbf{16.39×} and \textbf{27.49×} acceleration in the core- and ordinary-agent stages while cutting token consumption to \textbf{less than 20\%} of the baseline. Beyond efficiency, GASim achieves superior geometric alignment with real-world public opinion trends. We also validate GOM on the memory retrieval benchmark LoCoMo \cite{LOCOMO}, where it sets a new state-of-the-art with \textbf{71.56\%} accuracy.

Our contributions can be summarized as follows:
\begin{itemize}[itemsep=0.5pt, topsep=2pt]
    \item We introduce GASim, a graph-accelerated hybrid multi-agent framework for large-scale social simulations, with EDG as a hybrid coordinator that dynamically partitions agents into core and ordinary types.
    \item We propose GOM for core agents to rapidly retrieve memories with a lightweight graph-based memory model, alleviating the heavy latency in LLM-based retrieval process.
    \item We design GMP for ordinary agents to update opinion in parallel with fine-grained features and Graph Attention Network, resolving the sequential execution bottlenecks of ABMs.
\end{itemize}

\begin{figure*}[t] 
    \centering
    \includegraphics[width=\textwidth]{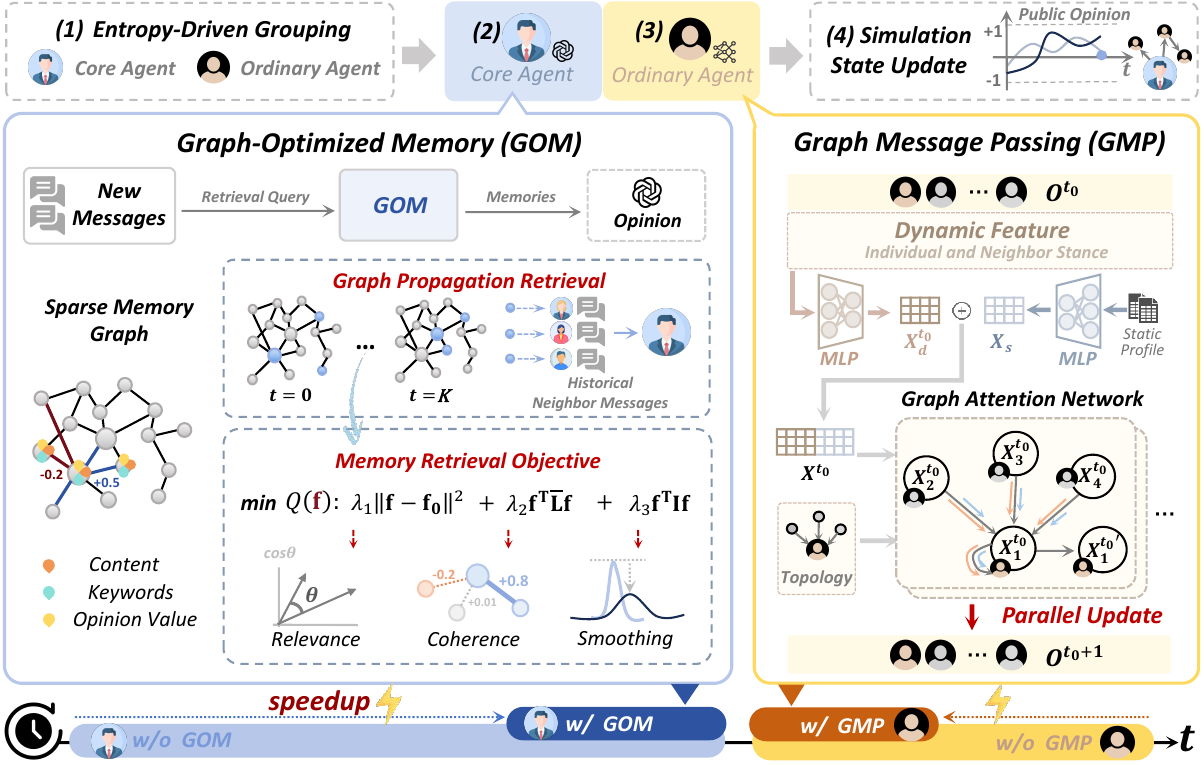} 
    \caption{Overall Pipeline of GASim. At each step, Entropy Driven Grouping (EDG) identifies emergent core and ordinary agents. Graph-Optimized Memory (GOM) accelerates LLM-driven core agents by updating a sparse memory graph and optimizing the retrieval vector $\mathbf{f}$ by a lightweight graph propagation. Graph Message Passing (GMP) updates agents’ opinion values in parallel via a Graph Attention Network with fine-grained features.}
    \label{fig:Overall Framework}
\end{figure*}

\section{Related Work}
\subsection{Costly Memory Design in Simulation}
Recent LLM-based multi-agent frameworks are highly expressive for large-scale social simulation, but suffer from severe latency due to memory designs that rely on heavy LLM-based processing. For example, keyword and contextual extraction in A-Mem \cite{A-mem}, memory rewriting in Mem0 \cite{mem0}, and importance or immediacy scoring in AgentVerse \cite{hisim} introduce substantial latency at scale. Vector similarity search methods such as FAISS \cite{FAISS} improve efficiency, but treat memories independently without explicit relationships, making it difficult to support coherent retrieval as memory grows.
To balance efficiency and accuracy, we propose GOM for LLM-driven core agents. It builds sparse memory graphs using similarity over content, keywords, and opinion values, and retrieves memories via lightweight graph propagation without LLM-generated edges.

\subsection{Latency in Agent-Based Models}
Traditional ABMs rely on sequential execution, leading to severe latency as the simulation scale increases \cite{HK, RA}. Recently, neural network–based ABMs have been proposed to parallelize agent updates and learn interaction dynamics \cite{fine-grained, gnnabm}. Although these models exhibit more limited reasoning ability compared to LLMs, they offer valuable insights for our hybrid acceleration paradigm: for the majority of ordinary agents representing the general crowd, we design GMP to parallelize agents' opinion updates by leveraging fine-grained opinion features and a Graph Attention Network, speeding up the simulation process.

\section{Proposed Framework: GASim}

The overall pipeline of GASim is illustrated in Figure~\ref{fig:Overall Framework}. GASim first employs an EDG module as a hybrid coordinator to distinguish core and ordinary agent types. Based on the grouping, a lightweight memory model GOM accelerates the core-agent stage, while a parallel updating model GMP efficiently speeds up the ordinary-agent stage.

\subsection{Entropy Driven Grouping}

To better model the dynamic emergence of opinion leaders, the EDG module is executed at the start of each simulation step. It dynamically identifies LLM-driven core agents, while the remaining agents function as general crowd agents driven by numerical models.

According to The People’s Choice \cite{PeopleChoice}, opinion leadership is an emergent behavior, not a fixed status. Traditional methods relying on static network degrees \cite{hisim} fail to capture this fluidity. In contrast, EDG employs dynamic entropy-based grouping to identify core agents who are embedded in information-diverse neighborhoods. 

Specifically, when categorizing agent $a_i$ for time step $t_0+1$, we compute the information entropy value $e_i^{t_0}$ of its neighborhood at time step $t_0$. Here, information entropy quantifies the diversity of opinions in an agent’s local environment, with higher values indicating broader viewpoint exposure,
\begin{equation}
\begin{aligned}
\label{information_entropy}
e_i^{t_0}&=-\sum\nolimits_{j} p_{ij}^{t_0}\cdot \text{log}_2p_{ij}^{t_0},   
\end{aligned}
\end{equation}
where $p_{ij}$ denotes the proportion of the $j$-th opinion value in agent $a_i$ neighborhood, and all $e_i^{t_0}$ form the entropy vector $\mathbf{e}^{t_0}$ to group agents. 
Following the Pareto principle \cite{pareto}, we select a small set of $K$ agents as core agents from the complete set $\mathcal{I}$ based on the information entropy,
\begin{equation}
\label{core_agents}
\mathcal{I}_c^{t_0+1}
= \left\{ a_i \in \mathcal{I} \;\middle|\; i \in \text{TopK}\left(\mathbf{e}^{t_0}\right) \right\},
\end{equation}
where $\text{TopK}(\cdot)$ returns indices with $K$ largest values. Then the ordinary agents can be denoted by $\mathcal{I}_o^{t_0+1} = \mathcal{I} - \mathcal{I}_c^{t_0+1}$. With this design, the model dynamically captures the evolving role of core agents as opinion leader, while the general crowd as ordinary agents.



\subsection{Graph-Optimized Memory}
\label{sec:gom}
The memory of core agents is sped up by GOM, which replaces costly LLM-based retrieval pipelines with a lightweight graph-based memory model. Unlike isolated similarity matching, the model encodes memory relations in a graph structure, amplifying coherent retrieval signals.

The complete behavior of a core agent is illustrated in Figure~\ref{fig:Overall Framework}. The agent first receives new messages from its neighbors. To simulate the human \textit{Observe--Recall--Act} decision cycle, agents must retrieve memories of previously observed neighbors' messages before performing personalized LLM-based social reasoning (e.g., posts, comments, retweets). Accordingly, we generate a high-level retrieval query $\mathbf{q}$ from the incoming neighbor messages, then GOM \textbf{a)} constructs a memory graph,  
\textbf{b)} formulates memory retrieval as an optimization problem, and \textbf{c)}
retrieves memories via a lightweight graph propagation algorithm. 

\paragraph{\textbf{a) Memory Graph Construction.}}
We encode the agent's memory as an opinion-aware, sparse weighted graph $G_{mem}=\left(\mathcal{V},\mathcal{E},\mathbf{W}\right)$, where $\mathcal{V}$, $\mathcal{E}$ and $\mathbf{W}$  respectively represent node set, edge set, and adjacency matrix. 
It prioritizes strongly opinionated memories, which are more influential than neutral experiences in shaping agent interactions in social simulations. Specifically, 
each node $\mathcal{V}_i$ represents a historical neighbor message with content $\mathbf{c}_i$, content embedding $\mathbf{m}_i$, keyword embedding $\mathbf{k}_{i}$ and an opinion value $o_i \in [-1, +1]$ where $-1$ denotes extremely negative and $+1$ denotes extremely supportive stances. 
Given $n$ memories, $\mathbf{W}\in \mathbb{R}^{n \times n}$ is a sparse symmetric adjacency matrix that connects each memory $i$ to its top-$k$ most similar memories, with the edge weight
\begin{equation}
\label{edge_weight}
w_{ij} = o_i \cdot o_j \cdot \cos(\mathbf{m}_i, \mathbf{m}_j).
\end{equation}
This graph design encodes stance and semantic consistency, prioritizing strongly opinionated memories over neutral ones.

\paragraph{\textbf{b) Retrieval Objective.}}
Rather than selecting memories independently based on query similarity, GOM treats the retrieval as a global optimization problem over the memory graph. Our goal is to optimize a retrieval probability vector $\mathbf{f}$, with element $f_i \in [0,1]$ denoting the retrieval probability of memory $i$. The retrieval objective is designed to balance three competing objectives:
\begin{itemize}[leftmargin=1.2em, itemsep=0.5pt, topsep=2pt]
    \item \textit{Relevance.} Maintain the initial similarity between the query and memories.
    \item \textit{Consistency.} Leverage the graph structure to encourage semantically and stance-consistent memories connected by high-weight edges to receive similar retrieval scores.
    \item \textit{Smoothing.} Penalize solutions that concentrate probability mass on only few memories, encouraging a smoother distribution.
\end{itemize}
Consequently, we formulate the objective as
\begin{equation}
\label{retrieve_function}
\textbf{min}\ Q(\mathbf{f})=\lambda_1 \|\mathbf{f}-\mathbf{f_0}\|^2+\lambda_2\mathbf{f}^\text{T} \overline{\mathbf{L}}\mathbf{f}+\lambda_3\mathbf{f}^\text{T}\mathbf{I}\mathbf{f},
\end{equation}
where $\lambda_{1}, \lambda_{2}, \lambda_{3}$ weight the respective terms:
\begin{itemize}[leftmargin=1.2em, itemsep=0.5pt, topsep=2pt]
    \item The first term anchors $\mathbf{f}$ to the initial relevance scores $\mathbf{f_0}$ based on the query $\textbf{q}$. It is defined as $(\mathbf{f_0})_i=\frac{1}{2}(\text{cos}(\textbf{q},\,\mathbf{m}_i)\,+ \text{H}_{\tau}(\text{cos}(\textbf{q},\,\textbf{k}_{i})))$, where $\text{H}_{\tau}(x)=1$ if $x \geq\tau$ and 0 otherwise. 
    \item The second term employs the normalized graph Laplacian $\overline{\mathbf{L}}=\mathbf{I}-\mathbf{D}^{-\frac{1}{2}}\mathbf{W}\mathbf{D}^{-\frac{1}{2}}$ to penalize differences between connected nodes, ensuring that memories with high edge weights receive similar scores. In this formulation, $\mathbf{I}$ is the identity matrix and $\mathbf{D}$ is the diagonal degree matrix of the graph, where $d_{ii}=\sum_{j}w_{ij}$.
    \item The third term regularizes the magnitude of $\mathbf{f}$ to encourage a smoother distribution.
\end{itemize}

Since $G_{mem}$ may contain negative weights ($o_io_j<0$), the objective can be non-convex and difficult to optimize, with more details in Appendix~\ref{appendix: GOM_closed_form}. While Softmax is typically used to map inputs to positive probabilities, it is unsuitable here as opinion values carry specific semantic interpretations mentioned in the memory graph construction. To ensure convexity, we introduce a correction term $\mathbf{\Delta}$, whose diagonal elements are defined as $\mathbf{\Delta}_{ii}=\nu(\,\sum_j \left| w_{ij}\right|-d_{ii} \,), \nu \geq 1$. We can then rewrite the objective function in \eqref{retrieve_function},
\begin{equation}
\begin{aligned}
\label{modified_retrieve_function}
\overline{\mathbf{L}^{\prime}}&=\mathbf{\overline{\mathbf{L}} +\mathbf{D}^{-\frac{1}{2} }\Delta\mathbf{D}^{-\frac{1}{2}}}, \\
\widetilde{Q}(\mathbf{f})=\lambda_1 &\|\mathbf{f}-\mathbf{f_0}\|^2 +\lambda_2\mathbf{f}^\text{T} \overline{\mathbf{L}^{\prime}}\mathbf{f} +\lambda_3\mathbf{f}^\text{T}\mathbf{I}\mathbf{f}.
\end{aligned}
\end{equation}
The convexity of $\widetilde{Q}(\mathbf{f})$ is proven in Appendix~\ref{appendix: GOM_convexity}. By letting $\nabla \widetilde{Q}(\mathbf{f})=0$, we can obtain the closed form $\mathbf{f}^*=\lambda_1\left[\,(\lambda_1+\lambda_3)\mathbf{I}+\lambda_2 \overline{\mathbf{L}^{\prime}} \,\right]^{-1}\mathbf{f_0}$, where each element $f_i^{*}$ represents the optimal retrieval possibility of memory $i$ under the query $\textbf{q}$, with the proof in Appendix~\ref{appendix: GOM_closed_form}.

\paragraph{\textbf{c) Graph Propagation Retrieval.}}
We note that explicitly computing $\mathbf{f}^*$ requires matrix inversion, which is prohibitively expensive for large-scale memory graphs. Therefore, we introduce a lightweight graph propagation algorithm to approximate the optimal retrieval vector $\mathbf{f}^*$ via the designed iterative update,
\begin{equation}
\label{iteration}
\mathbf{f}_{k+1}=\mu(-\overline{\mathbf{L}^{\prime}})\, \cdot \,\mathbf{f}_{k}+(1-\mu) \, \cdot {\mathbf{f_0^{\prime}}},
\end{equation}
where $\mu=\frac{\lambda_2}{1-\lambda_2+\lambda_3}$ and $\mathbf{f_0^{\prime}}=\frac{\lambda_1}{2\lambda_1+\lambda_3-1}\mathbf{f_0}$ under the assumption $\lambda_1+\lambda_2=1$. After $K$ iterations, the optimal retrieved set of $R$ memories is given by $\mathcal{C}=\{\text{c}_i \mid i \in \text{TopR}(\mathbf{f}_K)\}$. The detailed proof is provided in the Appendix~\ref{appendix: GOM_propagation}. This algorithm avoids inverting the large matrix in the closed-form $\mathbf{f}^*$, reducing complexity from $O(n^3)$ to $O(Knr)$, where $K$ is the iteration count, $n$ the number of memory nodes, $r$ the number of nonzero edges in the sparse graph ($K \ll n, \ r \ll n$). 

As a result, GOM significantly mitigates the latency of LLM-heavy memory frameworks while preserving high retrieval accuracy, enabling core agents to perform efficient, human-like memory retrieval without excessive computational overhead.

\subsection{Graph Message Passing}
\label{sec:gmp}
Traditional ABMs rely on sequential execution and global synchronization, leading to substantial latency at scale. In contrast, GMP speeds up ordinary agents by modeling the opinion evolution as parallel message passing on a social graph. 

To preserve the interpretability of rule-based ABMs, GMP integrates dynamic opinion states with static agent attributes, thereby capturing fine-grained interaction dynamics. This formulation naturally motivates the adoption of a Graph Attention Network (GAT) \cite{gat}, which adaptively weights neighbor influence through the attention mechanism via a fully parallelized forward pass.

The complete behavior of ordinary agents is illustrated in Figure~\ref{fig:Overall Framework}. At a given time step $t_0$, GMP takes the historical opinions of all agents to encode dynamic opinion features $\boldsymbol{X}_d^{t_0}$, which captures individual and neighborhood stances. These are then concatenated with static agent profile features $\boldsymbol{X}_s$ to form the unified representation $\boldsymbol{X}^{t_0}$. 
The GAT propagates these features over the agent interaction graph $\mathcal{E^{\prime}}$, 
simultaneously updating all agents’ opinions $\mathbf{o}^{t_0+1}$ in a single forward pass
\begin{equation}
\label{gat_reasoning}
    \boldsymbol{o}^{t_0+1} =\ f_{\textbf{GAT}}(\boldsymbol{X}^{t_0}, \mathcal{E^{\prime}}).
\end{equation}
The predicted opinions $\boldsymbol{o}^{t_0+1}$ are appended to the opinion histories, which serves as the recursive input to GMP for the subsequent time step. Training details are provided in the Appendix~\ref{appendix: GMP_training}. 

The following details the construction of the unified representation $\boldsymbol{X}^{t_0}$. For a population of $N$ agents at time step $t_0$, GMP organizes raw opinion histories into global tensors to enable parallel processing: a global opinion tensor $\boldsymbol{S}^{t_0}\in \mathbb{R}^{N \times {t_0}}$ containing all agents' opinion histories, and a global neighbor opinion tensor $\boldsymbol{N}^{t_0} \in \mathbb{R}^{N \times M \times {t_0}}$. $\boldsymbol{N^{t_0}}$ is constructed by gathering the corresponding history vectors from $\boldsymbol{S}^{t_0}$ based on the interaction topology, followed by padding to the maximum degree $M$. Based on these two tensors, we extract two groups of dynamic features via parallel operations, including the individual stance feature $\boldsymbol{\varphi}_I^{t_0}$ and the neighbor stance feature $\boldsymbol{\varphi}_C^{t_0}$,
\begin{equation}
\begin{aligned}
\label{dynamic_matrix}
    \boldsymbol{\varphi}_I^{t_0}&= \left[ \boldsymbol{\mu}^{t_0}, \boldsymbol{\sigma}^{t_0}, \boldsymbol{o}_\text{max}^{t_0}, \boldsymbol{o}_\text{min}^{t_0}, \boldsymbol{o}_\text{last}^{t_0} \right]_{\boldsymbol{S}^{t_0}} \ , \\
    \boldsymbol{\varphi}_C^{t_0} &= \left[ \boldsymbol{\hat{\mu}}^{t_0}, \boldsymbol{\hat{\sigma}}^{t_0}, \text{sim}^{t_0}, \text{ech}^{t_0} \right]_{\boldsymbol{N}^{t_0}} \ ,
\end{aligned}
\end{equation}
where detailed operations are provided in the Appendix~\ref{appendix: GMP_dynamic_feature}. Specifically, $\boldsymbol{\varphi}_I^{t_0} \in \mathbb{R}^{N \times 5} $ summarizes each agent's individual opinion history (mean, deviation, max, min and the lastest value) to characterize personal tendency. $\boldsymbol{\varphi}_C^{t_0}\in \mathbb{R}^{N \times 4}$ aggregates neighborhood statistics (mean, deviation, pearson correlation and echo-chamber score) to quantify collective pressure and homophily. We concatenate the two feature groups in formula \eqref{dynamic_matrix} to obtain the dynamic stance representation $\boldsymbol{\varphi}_d^{t_0} = \left[\boldsymbol{\varphi}_I^{t_0} \middle\| \boldsymbol{\varphi}_C^{t_0}\right] \in \mathbb{R}^{N \times 9}$. Meanwhile, to provide contextual background (e.g., values, identity, and interests) beyond observable opinion histories, we encode agents’ profile text with a BERT encoder to obtain static social attribute embeddings, denoted by $\boldsymbol{\varphi_s} = \text{bert} \left(\text{Tokenizer}(\textbf{text}) \right) \in \mathbb{R}^{N \times d_b}$. 

To learn higher-order, nonlinear interactions among agents, two multi-layer perceptrons (MLPs) are applied to perform nonlinear transformations on the dynamic and static features. The MLPs project $\boldsymbol{\varphi}_d^{t_0}$ and $\boldsymbol{\varphi}_s$ into a 
higher dimensional latent space,
\begin{equation}
\begin{aligned}
\label{feature_construction}
    \boldsymbol{X}_d^{t_0} = &\boldsymbol{W}_2\left(\text{ReLU} \left(\boldsymbol{W}_1 \boldsymbol{\varphi}_d^{t_0} + \boldsymbol{b}_1 \right) \right) + \boldsymbol{b}_2, \\
    \boldsymbol{X}_s = &\boldsymbol{\hat{W}}_2\left(\text{ReLU} \left(\boldsymbol{\hat{W}}_1 \boldsymbol{\varphi}_s + \boldsymbol{\hat{b}}_1 \right) \right) + \boldsymbol{\hat{b}}_2, \\
\end{aligned}
\end{equation}
where $\boldsymbol{W}$ and $\boldsymbol{b}$ represent learnable weights and biases. We then concatenate these features to form the unified representation $\boldsymbol{X}^{t_0} = \left[\boldsymbol{X}_d^{t_0} \ \middle\| \ \boldsymbol{X}_s \right]$. As shown in formula~\eqref{gat_reasoning}, this representation serves as the input to the GAT—along with the agent interaction graph—to update opinions in parallel.

In summary, to replace sequentially executed ABM, GMP adopts parallel feature aggregation with one-shot GAT updating. This accelerates simulation while modeling fine-grained inter-agent dependencies at scale. See Appendix~\ref{appendix: GMP} for details.

\subsection{Simulation State Update}
While LLM-driven core agents generate textual opinions via GOM-retrieved memories, ordinary agents directly output numeric opinion values based on GMP. To align these heterogeneous outputs, we map each core agent’s generated text to a scalar in $[-1,1]$ using an LLM-based scorer (see Appendix~\ref{appendix: dataset}). These quantified scores are then aggregated with the ordinary agents' values to update the simulated opinion trend curve. The details of agent heterogeneity designs are clarified in Appendix~\ref{appendix: heterogeneity}.

\section{Experiment}

\subsection{Datasets and Metrics}
\label{datasets and metrics}
To evaluate GASim, we constructed three topic-based datasets from popular social media platforms using open-source crawlers \textit{Apify} and \textit{WeiboSpider}, with statistical details reported in Table~\ref{tab:dataset}.
\begin{table}[htbp]
    \centering
    \small
    \renewcommand{\arraystretch}{1.3}
    {
    \setlength{\tabcolsep}{4pt} 
    \begin{tabular}{cccc}
        \hline
        \textbf{Dataset} & \textbf{Users} & \textbf{Tweets} & \textbf{Time Span} \\
        \hline
        Politics & 9,135 & 12,404 & May 10 - Dec 06, 2017 \\
        Business & 9,150 & 14,494 & March 01 - July 29, 2021 \\
        Education & 11,454 & 135,528 & June 13 - Nov 10, 2024 \\
        \hline
    \end{tabular}
    }
    \caption{Statistics of the dataset.}
    \label{tab:dataset}  
\end{table}

Specifically, each processed data contains anonymized user id, user description, follower count, following count, tweet content, posting time and an opinion value. The Politics dataset covers discussions on Donald Trump and alleged Russian interference in the 2016 U.S. election, and the Business dataset covers global controversies surrounding Xinjiang-produced cotton, both collected from X (formerly Twitter). The Education dataset covers Sina Weibo discussions on alleged cheating in the Alibaba Global Math Competition. For alignment, all datasets are standardized to 30 time steps. See Appendix~\ref{appendix: dataset} for details.


To evaluate simulation quality, we leverage four metrics comparing simulated and ground-truth public opinion curves from complementary perspectives. \textbf{Statistical metrics} assess macro-level trend consistency, where $\boldsymbol{\Delta{\textbf{Bias}}}$ measures the average deviation, $\boldsymbol{\Delta{\textbf{Div}}}$ captures the temporal stability,
and \textbf{Corr.} computes the Pearson correlation coefficient for linear alignment. \textbf{The Geometric metric} \textbf{F.} (Fréchet distance), measures global curve similarity. Detailed mathematical definitions of the above metrics are provided in Appendix~\ref{appendix: macro_metrics}.

\begin{table}[!t]
    \centering
    \small 
    \renewcommand{\arraystretch}{1.2} 
    
    \begin{tabular}{lccc}
    \toprule
    \textbf{Metric} & \textbf{HiSim} & \textbf{GASim} & \textbf{Speedup} \\
    \midrule
    $\mathbf{T_\text{core}}$  & 316.33 & \textbf{19.30} & \textbf{16.39}$\times$ \\
    $\mathbf{T_\text{ordi}}$  & 84.13  & \textbf{3.06}  & \textbf{27.49}$\times$ \\
    \midrule 
    $\mathbf{T_\text{total}}$ & 401.84 & \textbf{40.43} & \textbf{9.94}$\times$  \\
    \bottomrule
    \end{tabular}
    \caption{Latency Analysis (min), where $\mathbf{T_\text{core}}$, $\mathbf{T_\text{ordi}}$ and $\mathbf{T_\text{total}}$ denote the runtime of core, ordinary agents and the total simulation (including other overheads).}
    \label{tab:Latency_Ablation}
\end{table}

\begin{figure}[!t]
    \centering 
    \includegraphics[width=0.5\textwidth]{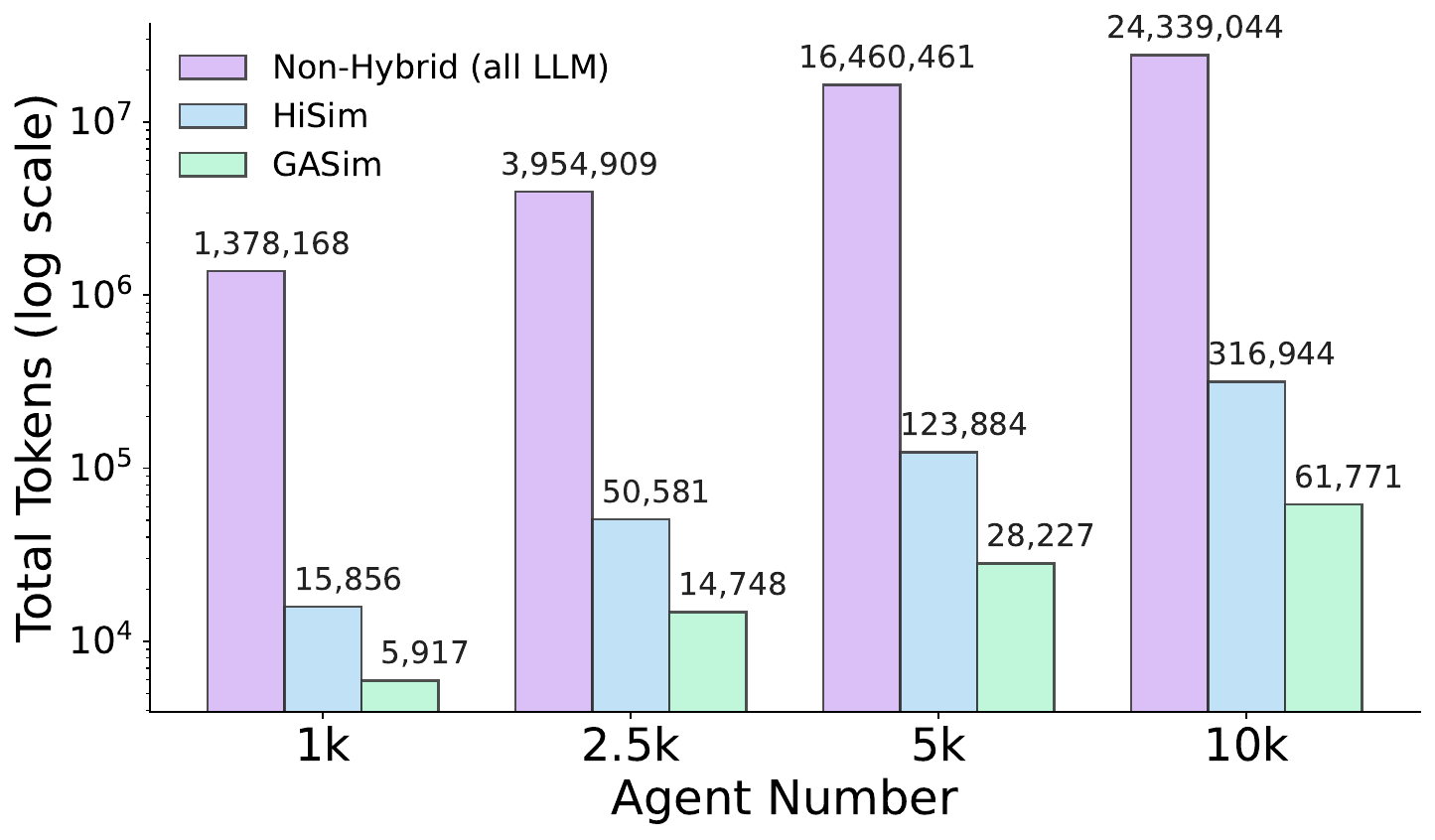} 
    \caption{Total token consumption across agent scales.} 
    \label{fig:total_token} 
\end{figure}

\begin{table*}[t]
    \centering
    \small
    \renewcommand{\arraystretch}{1.4} 
    \resizebox{\textwidth}{!}{
    \begin{tabular}{c|cccc|cccc|cccc}
    \hline
    \multirow{2}{*}{\textbf{Methods}} & \multicolumn{4}{c|}{\textbf{Politics}} & \multicolumn{4}{c|}{\textbf{Business}} & \multicolumn{4}{c}{\textbf{Education}} \\
     & $\boldsymbol{\Delta\textbf{Bias}} \downarrow$ & $\boldsymbol{\Delta\textbf{Div}} \downarrow $ & $\textbf{Corr.} \uparrow$ & $\textbf{F.} \downarrow$ & $\boldsymbol{\Delta\textbf{Bias}} \downarrow$ & $\boldsymbol{\Delta\textbf{Div} \downarrow} $ & $\textbf{Corr.} \uparrow$ & $\textbf{F.} \downarrow$ & $\boldsymbol{\Delta\textbf{Bias}} \downarrow$ & $\boldsymbol{\Delta\textbf{Div} }\downarrow $ & $\textbf{Corr.} \uparrow$ & $\textbf{F.} \downarrow$ \\
    \hline
    HK & 0.2003	& 0.0089 & 0.0581 & 0.3367 & 0.1081 & 0.0074 & 0.1214 & \underline{0.2369}  & 0.4828	& 0.0140 & \underline{0.6498} & 0.6293 \\
    RA  & 0.1629 & 0.0886 & \underline{0.2692} & 0.3346  & \underline{0.1046}	& \underline{0.0073} & \underline{0.4522} & 0.2438 & 0.4822	& \underline{0.0130}	& 0.2011 & 0.6242 \\
    Lorenz & 0.2339 & 0.1074 & -0.0637 & 0.4199  & 0.1298	& 0.0082 & -0.1228 & 0.2555	& 0.565 & 0.0199	& -0.3216 & 0.7579  \\
    SOD	& 0.1084 & \underline{0.0086}	& 0.1277 & 0.2464  & 0.1672 & 0.0105 & 0.068	& 0.3027 & 0.2716 & 0.0137 & 0.4013 & 0.3174   \\
    HiSim & \underline{0.1069} & 0.0167	& -0.003 & \underline{0.1622}  & 0.2302 & 0.0103	& -0.3532 & 0.3390 & \underline{0.2475} & 0.0167	& 0.388 & \underline{0.2237} \\
    GASim (Ours) &  \textbf{0.0700} &  \textbf{0.0074} &  \textbf{0.4261} &  \textbf{0.1349} &  \textbf{0.0807} &  \textbf{0.0060} &  \textbf{0.4707}	&  \textbf{0.1390} &	\textbf{0.0716}	& \textbf{0.0058} & \textbf{0.7686} &\textbf{0.1081} \\ [1pt]
    \hline           
    \end{tabular}
    }
    \caption{Trend alignment results in large-scale social simulation, where $\downarrow$ indicates that lower values are better, while $\uparrow$ indicates that higher is preferable. \textbf{Bold} and \underline{underline} indicate the best and second-best results.}
    \label{tab:10k_trend_loss}
\end{table*}

\subsection{Experimental Setting}

The agent population is set to 10,000, with the top-$K$ ($K=100$) agents selected as core type and driven by a locally deployed Llama-3.1-8B-Instruct (256 tokens, temperature 1). Bge-small-en-v1.5 is used to embed the memory contents. The GOM update parameter $\mu$ to 0.5 (i.e., $\lambda_i=0.5$, $1\leq i \leq 3$ ) to balance graph-based memory propagation and initialization, with $\nu=1$ in correction term $\mathbf{\Delta}$ and $\tau=0.9$ in $\mathbf{H}_{\tau}(x)$. All evaluation and scoring are performed using gpt-4o-mini API. Experiments run on a server with 40 vCPUs (Intel Xeon Platinum 8481C), two NVIDIA vGPU-48GB, and 180 GB RAM.

\subsection{Latency and Cost Analysis}
Table~\ref{tab:Latency_Ablation} demonstrates the acceleration performance of GASim on a large-scale simulation comprising 10,000 agents over 30 time steps. Compared to the traditional hybrid framework that relies on LLM-based memory retrieval and sequential ABMs, GASim achieves substantial efficiency gains. Specifically, adding GOM yields a \textbf{16.39$\times$} speedup (from 316.33 to 19.30 minutes) by replacing multi-stage LLM processing with lightweight graph-based memory retrieval. Similarly, integrating GMP accelerates the ordinary-agent stage by \textbf{27.49$\times$} (reducing time from 84.13 to 3.06 minutes) by parallelizing opinion updates via batched tensor processing in a single graph attention forward pass. Although $\mathbf{T_\text{total}}$ includes minor I/O overheads for persisting historical embeddings and keywords for driving GOM, this cost is negligible compared to the computational gains. Consequently, the overall simulation time is reduced from 6.69 to 0.67 hours, representing a significant \textbf{9.94$\times$ acceleration}.

In addition to faster simulation speeds, GASim dramatically lowers operational costs. As shown in Figure~\ref{fig:total_token}, GASim dramatically reduces total token consumption compared to HiSim and the non-hybrid framework, with the efficiency advantage increasing as the number of agents grows. At 10,000 agents, GASim consumes only 61,771 tokens, corresponding to approximately 1/5 of HiSim (316,944 tokens) and 1/400 of the non-hybrid baseline (24.3M tokens). This efficiency gain arises not only from the hybrid design but also from the introduction of GOM for core agents, which avoids invoking LLMs during memory storage and retrieval.

\subsection{Trend Alignment Evaluation}

To assess the fidelity of GASim, we evaluate trend alignment by comparing the simulated and real opinion curve using metrics in Section~\ref{datasets and metrics}. See Appendix~\ref{appendix: baseline_alignment} for baseline methods.

As detailed in Table~\ref{tab:10k_trend_loss}, GASim demonstrates superior performance across all metrics on diverse domain events. Regarding statistical metrics, GASim maintains a remarkably low $\boldsymbol{\Delta\textbf{Bias}}$, staying below 1\% across all domains. 
Meanwhile, GASim achieves an average reduction of 29.05\% in $\boldsymbol{\Delta\textbf{Div}}$ and an average improvement of 26.89\% in $\textbf{Corr.} $ compared to the second-best method. 
It also achieves the lowest F. across datasets, indicating superior geometric similarity.
\begin{figure}[h] 
    \centering 
    \includegraphics[width=0.48\textwidth]{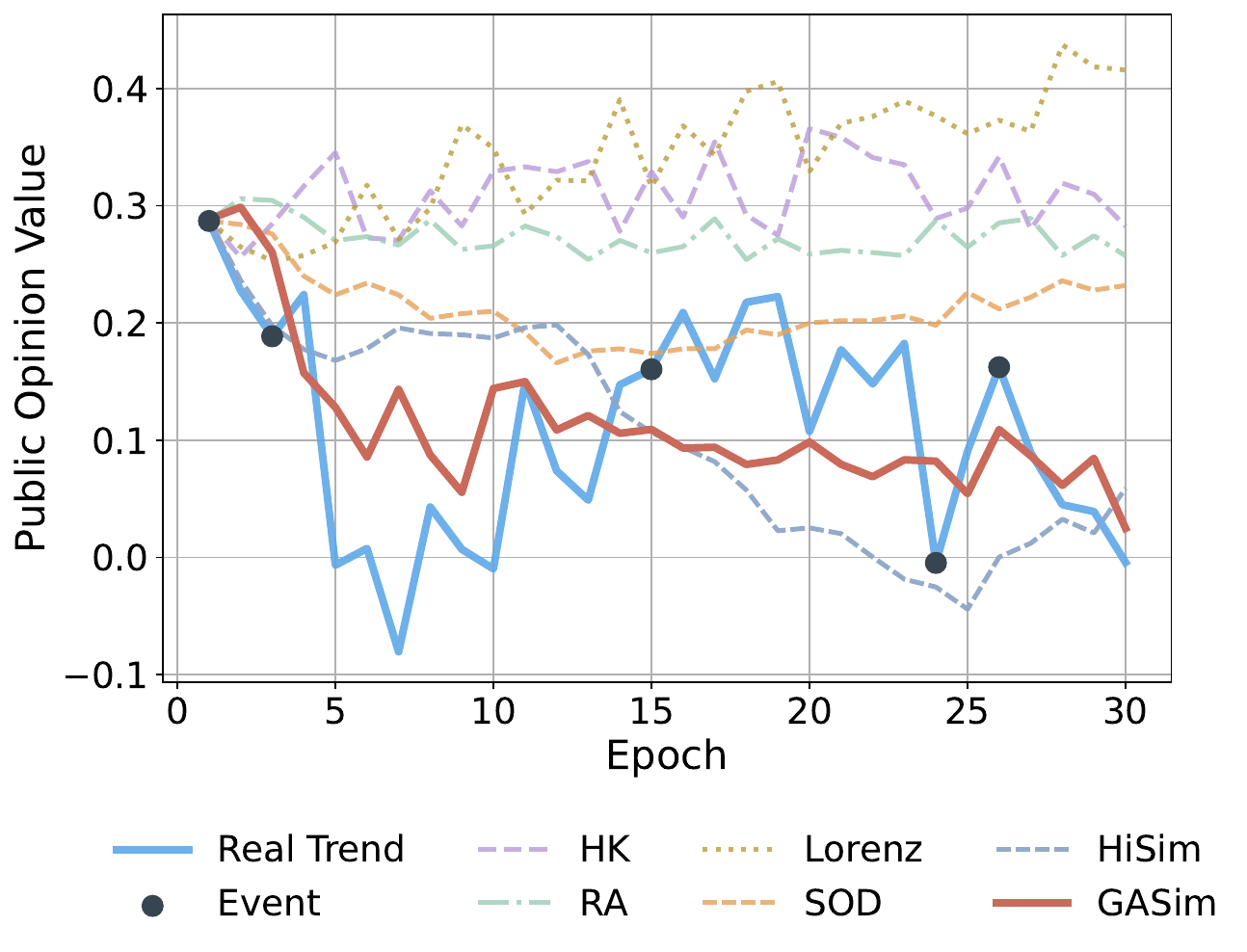} 
    \caption{Visualization of trend alignment results across different methods on the Politics dataset} 
    \label{fig:macro_trend} 
\end{figure}

\begin{table*}[t]
    \centering
    \scriptsize
    
    \renewcommand{\arraystretch}{1.3}
    \resizebox{\textwidth}{!}{
    \begin{tabular}{c|c|ccccc}
        \hline
        \multirow{2}{*}{\textbf{Method}}
        & \multirow{2}{*}{\makecell{\textbf{Chunk Size}}}
        & \multicolumn{5}{c}{\textbf{Accuracy (\%, LLM-as-a-Judge)}}\\
        & & \textbf{Single Hop} & \textbf{Multi-Hop} & \textbf{Open Domain} & \textbf{Temporal} & \textbf{Overall} \\[1pt]
        \hline
        A-Mem   & 2520 & $39.79 \pm 0.38$ & $18.85 \pm 0.31$ & $54.05 \pm 0.22$ & $49.91 \pm 0.31 $ & $48.38 \pm 0.15$ \\
        LangMem & 127 & $62.23 \pm 0.75$ & $47.92 \pm \ 0.47$ & $71.12 \pm \ 0.2$ & $23.43 \pm \ 0.39$ & $58.10 \pm \ 0.21$ \\
        Zep     & 3911 & $61.7 \pm 0.32$  & $41.35 \pm 0.48$ & $\mathbf{76.6 \pm 0.13}$  & $49.31 \pm 0.50$ & $65.99 \pm 0.16$  \\
        Mem0    & 1764 & \underline{$67.13 \pm 0.65$} & $\underline{51.15 \pm 0.31}$ & $72.93 \pm 0.11$ & $55.51 \pm 0.34$& $66.8 \pm 0.15$   \\ 
        Mem0g   & 3616 & $65.71 \pm 0.45$ & $47.19 \pm 0.67$ & \underline{$75.71 \pm 0.21$} & $\underline{58.13 \pm 0.44} $ & $68.44 \pm 0.17$ \\[1pt]
        GOM (Ours) & 2492 & $\mathbf{75.39 \pm 0.51}$ & $\mathbf{59.6 \pm 0.53}$  & $74.96 \pm 0.18$ &  $\mathbf{70.7 \pm 0.47}$ & $ \mathbf{71.56 \pm 0.20}$ \\
        \hline
    \end{tabular}}
    \caption{Evaluation of memory retrieval performance: GOM vs. baselines on LoCoMo dataset \cite{LOCOMO}, where Chunk Size indicates the average context length, including retrieved contents and the answer prompt.}
    \label{tab:LoCoMo_test}
\end{table*}

To provide a qualitative comparison, Figure~\ref{fig:macro_trend} visualizes simulation results on the Politics dataset.
Consistent with Table~\ref{tab:10k_trend_loss}, GASim produces a trajectory that closely aligns with the real-world opinion trend, whereas baseline methods fail to capture opinion evolution over long durations.
Specifically, traditional ABMs (HK, RA, and Lorenz) tend to fluctuate merely around the initial opinion value, constrained by their rigid and predefined mathematical update rules. 
Among LLM-based methods, SOD better approximates real trends than ABMs due to semantic reasoning and agent bias design, but its random one-to-one communication limits its ability to capture dynamic opinion fluctuations. In contrast, the hybrid framework HiSim exhibits extreme and one-sided shifts caused by static degree-based agent grouping. GASim outperforms both by leveraging GOM for accurate memory retrieval, GMP for fine-grained neighbor-aware opinion reasoning, EDG for dynamically coordinating agents via entropy-based partitioning to model emergent opinion leadership.

\subsection{Memory Architecture Evaluation}

We further evaluate GOM on the LoCoMo dataset \cite{LOCOMO}, a benchmark for long-term conversational memory. LoCoMo contains 10 long conversations, with question–answer pairs spanning single-hop, multi-hop, temporal and open-domain categories. For the evaluation, instead of using F1 and BLEU that rely on lexical overlap and often fail to reflect factual correctness, we adopt LLM-as-a-Judge to verify answer accuracy, assessing semantic and factual quality in a way that better aligns with human judgment. The judge prompt is provided in Appendix~\ref{appendix: memory_judge}.

As shown in Table~\ref{tab:LoCoMo_test}, GOM achieves state-of-the-art performance compared to a suite of competitive memory-based baselines, reaching an overall accuracy of 71.56\%. In particular, GOM shows substantial improvements on Single-Hop, Multi-Hop, and Temporal question types, outperforming the strongest existing models by approximately 10\%. These three categories respectively require locating a single factual span within one dialogue turn, synthesizing information dispersed across multiple conversation sessions, and accurate modeling of event sequences and their temporal ordering, all of which directly benefit from GOM’s graph-guided memory retrieval mechanism. In GOM, by formulating memory selection as a convex optimization problem over a structured memory graph, it is able to jointly consider semantic relevance, stance coherence, and smoothness of memory retrieval, rather than relying on isolated similarity matching.

While GOM underperforms on Open-Domain questions, this category primarily favors LLM-based memory baselines that leverage extensive prior knowledge. However, in social simulation, an over-reliance on such generic external knowledge can homogenize agent responses, thereby reducing personality diversity. Given that these questions constitute only 5\% of the dataset, the impact on overall results is negligible.

\begin{table}[t]
    \centering
    \scriptsize
    \renewcommand{\arraystretch}{1.2} 
    \resizebox{\columnwidth}{!}{
    \begin{tabular}{l|cccc}
    \hline
    {\textbf{Models}} & $\boldsymbol{\Delta\textbf{Bias}} \downarrow$ & $\boldsymbol{\Delta\textbf{Div}} \downarrow $ & $\textbf{Corr.} \uparrow$ & $\textbf{F.} \downarrow$ \\
    \hline
    GASim & \textbf{0.0700} & \textbf{0.0074} & \textbf{0.4261} & \textbf{0.1349} \\
    w/o GOM & 0.0771 & 0.0089 & 0.2942 & 0.1406 \\
    w/o GMP& 0.1027 & 0.1346 & -0.0989 &  0.2291 \\
    w/o EDG & 0.0872 & 0.0109 & 0.2528 & 0.1391 \\
    \hline           
    \end{tabular}
    }
    \caption{Ablation study of trend alignment on Politics dataset, where "w/o" denotes the removal of the module.}
    \label{tab:Accuracy_Ablation}
\end{table}

\subsection{Ablation Study}
To evaluate the specific contributions of each module to simulation fidelity, we conducted the ablation study focusing on public opinion trend alignment. As shown in Table~\ref{tab:Accuracy_Ablation}, removing any component leads to performance degradation across all metrics. 

Specifically, we first remove GOM by replacing our designed memory mechanism with HiSim's \cite{hisim} LLM-in-the-loop memory scheme, where the memory retrieval is guided by LLM-based assessments of importance and urgency. Under this setting, $\boldsymbol{\Delta\textbf{Bias}}$ increases to 0.0771 (a 10\% rise), $\boldsymbol{\Delta\textbf{Div}}$ increases to 0.0089 (a 20\% rise), and $\textbf{Corr.}$ drops to 0.2942 (a 30.96\% reduction). These results confirm that GOM's lightweight graph propagation is an effective alternative for agent's memory retrieval compared to computationally expensive multi-stage LLM processing. We then remove GMP by substituting the learned opinion update mechanism with a traditional agent-based model (Lorenz model by default), which causes significant deterioration across all metrics. This underscores the necessity of modeling the general crowd via fine-grained neural networks trained on real dynamic opinion data, rather than relying solely on LLMs which may be restricted by inherent training data biases. We also remove EDG by reverting the system to a static degree-based core assignment strategy, where core agents are selected solely based on in-degree (number of followers) rather than dynamic entropy-driven criteria. This results in a 47.3\% increase in $\boldsymbol{\Delta\textbf{Div}}$ and highlights the critical role of EDG in dynamically coordinating agents to stabilize the simulation's performance. Moreover, we conduct an additional experiment that verifies EDG-selected core agents correspond to empirically influential users. The experiment details are provided in Appendix~\ref{appendix: EDG}.

\section{Conclusion}
This paper introduces GASim, a graph-accelerated hybrid multi-agent framework for large-scale social simulations. By designing a lightweight graph-based memory model for core agents and a parallelized fine-grained neural network module for ordinary agents, GASim effectively addresses the latency challenges faced by  traditional hybrid frameworks. Extensive experiments confirm both the efficiency and accuracy of our framework. We anticipate that GASim will offer a new perspective on meeting low-latency requirements in the social simulation community.

\section*{Limitations}
Despite the effectiveness of GASim, our work has two key limitations. (1) The LLM-generated text lacks authenticity, and synthetic opinion value labels may reflect the inherent biases of the LLM itself. (2) Our simulation primarily focuses on textual interactions, while ignoring multimodal information (e.g., images and videos) that may also play a crucial role in public opinion dynamics.

\section*{Ethical Considerations}
GASim is designed as a research instrument for analyzing offline social dynamics rather than for active intervention. We acknowledge the potential dual-use risk associated with misuse in generating or amplifying harmful content. To mitigate this risk, all simulations are conducted in isolated and offline sandboxed environments with no connectivity to real-world platforms. 

Regarding data usage, we adhere to the following protocols: \textbf{(1) Privacy \& Content:} We pseudo-anonymize user identities by mapping public IDs to numerical indices. While the datasets contain controversial topics, such content is retained strictly to maintain the statistical fidelity of opinion dynamics. \textbf{(2) Consent:} Due to the dataset scale, obtaining individual consent was infeasible. We utilized only voluntarily public data in compliance with platform terms. \textbf{(3) IRB Status:} As this study analyzes existing public data without direct human intervention, it is exempt from formal IRB review.

\section*{Acknowledgments}
This work was supported by the National Key Research and Development Program of China (NO. 2024YFE0203200), the National Nature Science Foundation of China (NO. U24A20329, NO. 62527810), the Fundamental and Interdisciplinary Disciplines Breakthrough Plan of the Ministry of Education of China (NO. JYB2025XDXM103), and the Science Fund for Creative Research Groups (No.62121002).


\bibliographystyle{acl_natbib}
\bibliography{main}

\clearpage

\appendix

\section{Appendix}
\label{sec:appendix}

\subsection{Theoretical Analysis of GOM}
\label{appendix: GOM}

\subsubsection{Convexity Discussion}
\label{appendix: GOM_convexity}

In this section, we discuss the convexity of the original objective in formula \eqref{retrieve_function} and justify the need for the modification in formula \eqref{modified_retrieve_function} to achieve a more tractable optimization.

According to formula \eqref{retrieve_function}, it is straightforward to verify that the first and the third term are both convex function, since their Hessian matrices are equal to $\lambda_i \cdot 2 \mathbf{I} \succ 0$ $(i=1,3)$. Yet the second term of equation \eqref{retrieve_function} is not intuitive to judge its convexity. Motivated by \cite{NewsVerify}, we can rewrite the second term by defining $\overline{\mathbf{f}}=\mathbf{D}^{-\frac{1}{2}}{\mathbf{f}}$ as follows,
\begin{equation}
\begin{aligned}
\mathbf{f}^\text{T}\overline{\mathbf{L}}\mathbf{f}
  &= \mathbf{f}^\text{T}\left( \mathbf{I} - \mathbf{D}^{-\frac{1}{2}}\mathbf{W}\mathbf{D}^{-\frac{1}{2}} \right)\mathbf{f}\notag\\
  = \frac{1}{2}&\left(2 \overline{\mathbf{f}}^\text{T}\mathbf{D}\overline{{\mathbf{f}}}-2\overline{\mathbf{f}}^\text{T}\mathbf{W}\overline{\mathbf{f}}^\text{T} \right)\notag\\
  = \frac{1}{2}&\left(\sum_{i=1}^nd_{ii}\overline{f_i}^2 + \sum_{j=1}^{n}d_{jj}\overline{f_j}^2- 2\sum_{i,j=1}^{n} \overline{f_i} w_{ij}\overline{f_j}\right)\notag\\
  = \frac{1}{2}&\sum_{i, j=1}^{n} w_{ij}\left(\frac{f_i}{\sqrt{d_{ii}}}-\frac{f_j}{\sqrt{d_{jj}}}\right)^2.
\label{Equivalent}
\end{aligned}
\end{equation}
When we review the formula derivation in \eqref{Equivalent}, we can analyze that $\mathbf{f}^\text{T}\overline{\mathbf{L}}\mathbf{f}$ is convex when each of the edge weight $w_{ij} \geq 0$. In this case, $\mathbf{f}^\text{T}\overline{\mathbf{L}}\mathbf{f} \geq 0$, $\overline{\mathbf{L}}$ is positive semidefinite, and the whole term is convex with the minimal value being 0 when $\mathbf{f} \propto \sqrt{\mathbf{d}}=\left(\sqrt{d_{11}}, \sqrt{d_{22}}, \ldots, \sqrt{d_{nn}}\right)^\text{T}$. 

However, based on the sparse memory graph defined in section \textbf{3.2.1}, the memory edge weight $w_{ij}$ can be negative (since agents may have opposite opinion values), and the convexity of $\mathbf{f}^\text{T}\overline{\mathbf{L}}\mathbf{f}$ is no longer certain. Common approximations in existing research \cite{NewsVerify}\cite{Goldberg}, which model the processed data on a graph that contains negative edge weight, often forces $w_{ij} \geq 0$ by taking the absolute value $|w_{ij}|$. Yet these methods totally discard the opposite features in the data, which causes a large change in the normalized Laplacian matrix $\overline{\mathbf{L}}$. 

Therefore, to ensure the objective convexity for a more tractable optimization (the convexity of $\widetilde{Q}(\mathbf{f})$), we need to approximate $\overline{\mathbf{L}}$ by an appropriate modification. Based on the formula \eqref{modified_retrieve_function}, we introduce a correction term $\mathbf{\Delta}$, whose diagonal elements are defined as $\mathbf{\Delta}_{ii}=\nu(\,\sum_j \left| w_{ij}\right|-d_{ii} \,), \nu \geq 1$. By adding $\mathbf{D}^{-\frac{1}{2}} \mathbf{\Delta} \mathbf{D}^{-\frac{1}{2}}$ to the normalized Laplacian matrix $\overline{\mathbf{L}}$, we obtain a convex $\overline{\mathbf{L}^{\prime}}$, as shown in the proof below.
 
 \textbf{Proof}. Based on the Gershgorin's circle theorem, there exists an upper and lower bound for each eigenvalue $\lambda_i(\mathbf{L})$ of unnormalized $\mathbf{L}=\mathbf{D}-\mathbf{W}$, 
\begin{equation}
\begin{aligned}
\label{gershgorin}
\mathbf{L}_{ii}-\sum_j \lvert \mathbf{L}_{ij}\rvert 
&\leq \lambda_i(\mathbf{L})
\leq \mathbf{L}_{ii}+\sum_j \lvert \mathbf{L}_{ij}\rvert  \\
\quad d_{ii}-\sum_j \lvert w_{ij}\rvert 
&\leq  \lambda_i(\mathbf{L})
\leq d_{ii}+\sum_j \lvert w_{ij}\rvert  \\
\quad \sum_j (w_{ij}-\lvert w_{ij}\rvert) 
&\leq 
 \lambda_i(\mathbf{L})  
\leq \sum_j (w_{ij}+\lvert w_{ij}\rvert).
\end{aligned}
\end{equation}
Based on formula~\eqref{gershgorin}, if the memory graph contains negative edge weights $ w_{ij} < 0$, then the lower bound of the $i$-th eigenvalue of $\mathbf{L}$ falls below zero and $\mathbf{L}$ may not be positive semidefinite . However, with the diagonal matrix $\mathbf{\Delta}$ added on $\mathbf{L}$, the inequality \label{eigenvalue} becomes equation \eqref{modified_eigenvalue} 
\begin{equation}
 \begin{aligned}
 \label{modified_eigenvalue}
\left(\nu-1\right)\left(\sum_j \lvert w_{ij}\rvert-\sum_jw_{ij}\right) 
\leq \lambda_i(\mathbf{L+\Delta})\\  
\leq \sum_j w_{ij}+\sum_j \lvert w_{ij}\rvert + \nu\left(\sum_j \lvert w_{ij} \rvert - w_{ij} \right).
\end{aligned}
\end{equation}
Since $\nu \geq 1$, the lower bounds for all eigenvalues are greater than 0, which is $\{\lambda_i(\mathbf{L}) \}_{i=1}^n \in \mathbb{R}_+$. In this way, the matrix $\mathbf{L}+ \mathbf{\Delta}$ must be positive semidefinite. By normalization, we can obtain the convex 
$\overline{\mathbf{L}^{\prime}}=\overline{\mathbf{L}} + \mathbf{D}^{-\frac{1}{2}} \mathbf{\Delta} \mathbf{D}^{-\frac{1}{2}}$. With this approximation, only the diagonal elements of $\mathbf{L}$ are modified, ensuring the convexity of $\widetilde{Q}(\mathbf{f})$. This completes the proof. \(\square\)

\subsubsection{Closed-Form Solution}
\label{appendix: GOM_closed_form}

The closed-form optimal solution of formula \eqref{modified_retrieve_function} can be derived directly based on the formulation, which characterizes the retrieval probabilities of the memory nodes during the agent’s memory retrieval task. Since we have proven the convexity of equation \eqref{modified_retrieve_function}, the optimal $\mathbf{f}^*$ can be obtained by solving the first-order condition $\nabla \widetilde{Q}(\mathbf{f})=0$. Specifically, we can expand $\widetilde{Q}$ as $\lambda_1 \left(\mathbf{f}-\mathbf{f_0}\right)^\text{T}\left(\mathbf{f}-\mathbf{f_0}\right)+\lambda_2 \mathbf{f}^\text{T} \overline{\mathbf{L}^{\prime}}\mathbf{f} + \lambda_3 \mathbf{f}^\text{T} \mathbf{f}$. Then we can calculate the gradient of $\widetilde{Q}(\mathbf{f})$
\begin{equation}
\begin{aligned}
\label{gradient}
\nabla\widetilde{Q}(\mathbf{f})
&=\lambda_1  2\left(\mathbf{f}-\mathbf{f_0}\right) + \lambda_2 \left(\overline{\mathbf{L}^{\prime}}+{\overline{\mathbf{L}^{\prime}}}^\text{T}\right)\mathbf{f}+2\lambda_3\mathbf{f}\\
&=\left(\lambda_1+\lambda_3 \right)\cdot2\mathbf{f}+\lambda_2 \cdot 2\overline{\mathbf{L}^{\prime}}\mathbf{f}-\lambda_1 \cdot 2\mathbf{f}_0\\
&=\left[\,(\lambda_1+\lambda_2) \cdot \mathbf{I}\,+\lambda_2 \overline{\mathbf{L}^{\prime}}\right]\cdot2\mathbf{f}-\lambda_1 \cdot 2 \mathbf{f_0}.
\end{aligned}
\end{equation}
Let $\nabla \widetilde{Q}(\mathbf{f})=0$, and we can obtain the optimal memory selection vector $\mathbf{f}^*$ as follows
\begin{equation}
\label{closed-form}
\mathbf{f}^*=\lambda_1\left[\,(\lambda_1+\lambda_3)\mathbf{I}+\lambda_2 \overline{\mathbf{L}^{\prime}} \,\right]^{-1}\mathbf{f_0}.
\end{equation}
This completes the proof.  \(\square\)

\subsubsection{Proof of Graph Propagation Retrieval}
\label{appendix: GOM_propagation}

To reduce the computational complexity of calculating the 
closed-form of the retrieval vector $\mathbf{f}$, we design a fast graph propagation retrieval that avoids calculating the inverse of the large-scale matrix. 

For convenience, we assume the two hyperparameters $\lambda_1$ and $\lambda_2$ are set to satisfy $\lambda_1+\lambda_2=1$, then formula \eqref{closed-form} can be rewritten as formula \eqref{rewritten closed-form}
\begin{equation}
\begin{aligned}
\label{rewritten closed-form}
\mathbf{f}^*
&=(1-\lambda_2)\left[\,(1-\lambda_2+\lambda_3)\mathbf{I}+\lambda_2 \overline{\mathbf{L}^{\prime}} \,\right]^{-1}\mathbf{f_0} \\
&=\frac{1-\lambda_2}{1-\lambda_2+\lambda_3}\left[\mathbf{I}+\frac{\lambda_2}{1-\lambda_2+\lambda_3}\overline{\mathbf{L}^{\prime}}\right]^{-1}\mathbf{f_0} \\
&=\frac{\lambda_1}{2\lambda_1+\lambda_3-1} \cdot \left(1-\mu\right)\left[\mathbf{I}+\mu\overline{\mathbf{L}^{\prime}}\right]^{-1}\mathbf{f_0},\\
&=\left(1-\mu\right)\left[\mathbf{I}+\mu\overline{\mathbf{L}^{\prime}}\right]^{-1}\mathbf{f_0^{\prime}},
\end{aligned}
\end{equation}
where $\mu=\frac{\lambda_2}{1-\lambda_2+\lambda_3},\ \mathbf{f_0^{\prime}}=\frac{\lambda_1}{2\lambda_1+\lambda_3-1}\mathbf{f_0}$. 
Based on this simplification, we can efficiently calculate the optimal memory selection vector $\mathbf{f}^*$ in formula \eqref{rewritten closed-form} by the designed graph propagation retrieval $\mathbf{f}_{k+1}=\mu(-\overline{\mathbf{L}^{\prime}})\, \cdot \,\mathbf{f}_{k}+(1-\mu) \, \cdot {\mathbf{f_0^{\prime}}}$, with the proof detailed as follows.

\textbf{Proof}. Based on the initial value $\mathbf{f_0^{\prime}}$, we can make the following deduction in the iteration,
\begin{equation}
\begin{aligned}
\label{iteration_proof}
&\mathbf{f}_1 = \mu(-\mathbf{\overline{L^{'}}}) \cdot \mathbf{f_0^{\prime}} + (1-\mu) \cdot  \mathbf{f_0^{\prime}}, \\
&\mathbf{f}_2 = \left[ \mu(-\mathbf{\overline{L^{'}}}) \right]^2\cdot \mathbf{f_0^{\prime}} + (1-\mu)\left[\mu(-\mathbf{\overline{L^{'}}})+\mathbf{I} \right] \cdot  \mathbf{f_0^{\prime}},\\
&\hspace{10em} \vdots \\
&\mathbf{f}_k = \left[ \mu(-\mathbf{\overline{L^{'}}}) \right]^k\cdot \mathbf{f_0^{\prime}} + (1-\mu) \sum_{i=0}^{k-1}\left[\mu(-\mathbf{\overline{L^{'}}})\right]^{i} \cdot \mathbf{f_0^{\prime}}, \\
& \sum_{i=0}^{k-1}\left[\mu(-\mathbf{\overline{L^{'}}})\right]^{i} = \left(\mathbf{I}-\left[\mu(-\mathbf{\overline{L^{'}}})\right]^k \right)\left( \mathbf{I}+\mu(\mathbf{\overline{L^{'}}}) \right)^{-1}.
\end{aligned}
\end{equation}
Since our constructed memory graph $G_{mem}$ is sparse, the modified Laplacian matrix $\mathbf{\overline{L^{'}}}$ can be approximated to a sparse random matrix. Therefore, we can make the following derivation
\begin{equation}
\begin{aligned}
\label{convergence}
&\lim_{k \to \infty} \left[\mu(-\overline{\mathbf{L}^{'}}) \right]^k=0, \\
&\lim_{k \to \infty} \mathbf{f}_k = (1 - \mu)\left( \mathbf{I} + \mu\overline{\mathbf{L}^{'}} \right)^{-1} \mathbf{f_0^{\prime}},
\end{aligned}
\end{equation}
which is exactly the same as the equation form in \eqref{rewritten closed-form}. 
Therefore, the optimal memory selection vector $\mathbf{f}^*$ can be efficiently approximated using the proposed graph propagation retrieval algorithm, which avoids computing the inverse of the large-scale matrix. This completes the proof. \(\square\)



\subsection{Detailed Design of GMP}
\label{appendix: GMP}

In this section, we provide the specific architectural details of the Graph Message Passing (GMP) module. GMP takes the historical opinion values of all agents as input, and updates agents' opinions simultaneously in a single forward pass. Based on the method, the following sections detail the dynamic feature extraction, the feature projection, the graph attention layers, and the training design.

\subsubsection{Dynamic Feature Extraction}
\label{appendix: GMP_dynamic_feature}

To efficiently process the population of $N$ agents over accumulating time steps, we vectorize GMP feature extraction process. By organizing historical data into global tensors, we compute the raw dynamic features, including the individual stance features $\boldsymbol{\varphi}_I$ and neighbor stance features $\boldsymbol{\varphi}_C$, using matrix operations.

\paragraph{Initialization and Global Tensors}
Let $t_0$ be the current time step and $M$ be the maximum neighbor degree in agents' topology. We define three primary tensors to facilitate parallel computation:
\begin{itemize}
    \item \textit{Global opinion tensor} $\mathbf{S}^{t_0} \in \mathbb{R}^{N \times t_0}$: Each row $i$ contains historical opinion values of agent $i$.
    \item \textit{Global neighbor opinion tensor} $\mathbf{N}^{t_0} \in \mathbb{R}^{N \times M \times t_0}$: A 3D tensor where $\mathbf{N}_{i,j,:}$ represents the opinion history of the $j$-th neighbor of agent $i$, padded with zeros for nodes with degree $< M$.
    \item \textit{Neighbor Mask} $\mathbf{M} \in \{0, 1\}^{N \times M}$: A binary matrix where $\mathbf{M}_{i,j} = 1$ indicates a valid neighbor and $0$ indicates padding.
\end{itemize}

\paragraph{Vectorized Operations}
The feature extraction logic is summarized in Table~\ref{tab:dynamic_feature_ops}. Operations are performed along specific tensor dimensions to exploit GPU parallelism.

\begin{table*}[h]
\centering
\small
\renewcommand{\arraystretch}{1.8}
\begin{tabular}{l l l}
\toprule
\textbf{Feature Group} & \textbf{Feature Description} & \textbf{Vectorized Mathematical Operation} \\
\midrule
\multirow{4}{*}{\shortstack[l]{\textbf{Individual Stance}\\ $\boldsymbol{\varphi}_I \in \mathbb{R}^{N \times 5}$}} 
& Mean \& Variance & $\boldsymbol{\mu} = \operatorname{mean}_{t}(\mathbf{S}^{t_0}), \quad \boldsymbol{\sigma}^2 = \operatorname{var}_{t}(\mathbf{S}^{t_0})$ \\
& Boundary Values & $\mathbf{o}_{\text{max}} = \max_{t}(\mathbf{S}^{t_0}), \quad \mathbf{o}_{\text{min}} = \min_{t}(\mathbf{S}^{t_0})$ \\
& Current State & $\mathbf{o}_{\text{last}} = \mathbf{S}^{t_0}_{:, t_0}$ \\
\midrule
\multirow{5}{*}{\shortstack[l]{\textbf{Neighbor Stance}\\ $\boldsymbol{\varphi}_C \in \mathbb{R}^{N \times 4}$}} 
& Neighbor Mean & $\hat{\boldsymbol{\mu}} = \operatorname{mean}_{m, t}(\mathbf{N}^{t_0} \odot \mathbf{M})$ \\
& Neighbor Std. & $\hat{\boldsymbol{\sigma}} = \sqrt{\sum_{m,t} ((\mathbf{N}^{t_0} - \hat{\boldsymbol{\mu}})^2 \odot \mathbf{M}) \oslash \sum_{m} \mathbf{M}}$ \\
& Pearson Alignment & $\operatorname{sim}_i = \frac{1}{\sum_j \mathbf{M}_{i,j}} \sum_{j=1}^M \operatorname{Pearson}(\mathbf{S}^{t_0}_{i,:}, \mathbf{N}_{i,j,:}^{t_0}) \cdot \mathbf{M}_{i,j}$ \\
& Echo Chamber & $\text{ech} = \operatorname{sim} \oslash (1 + \hat{\boldsymbol{\sigma}})$ \\
\bottomrule
\end{tabular}
\caption{Parallel operations for dynamic feature extraction, broadcasted over the agent dimension $N$.}
\label{tab:dynamic_feature_ops}
\end{table*}

These operations yield the final dynamic feature tensor $\boldsymbol{\varphi}_d^{t_0} = [\boldsymbol{\varphi}_I^{t_0} \parallel \boldsymbol{\varphi}_C^{t_0}] \in \mathbb{R}^{N \times 9}$, where $\parallel$ denotes the concatenation operator.

\subsubsection{Feature Projection}
\label{appendix: GMP_feature_projection}
Before entering the graph network, the raw features are projected into a shared latent space to facilitate fusion. As described in formula~\ref{feature_construction}, we employ two separate Multi-Layer Perceptrons (MLPs) for this purpose:
\begin{itemize}
    \item \textit{Dynamic MLP:} Projects the dynamic opinion statistics $\boldsymbol{\varphi}_d^{t_0} \in \mathbb{R}^9$ to a latent vector of size $64$.
    \item \textit{Static MLP:} Projects the BERT-encoded profile embeddings $\boldsymbol{\varphi}_s \in \mathbb{R}^{768}$ to a latent vector of size $64$.
\end{itemize}
These projected vectors are concatenated to form the input node features $\boldsymbol{X}^{t_0} \in \mathbb{R}^{128}$ for the GAT.

\subsubsection{Graph Attention Layers}
\label{appendix: GMP_graph_attention}
The main reasoning module consists of a two-layer GAT configuration. We utilize edge weights within the attention mechanism to allow structural interaction strengths to modulate message passing.

\paragraph{Layer 1 (Multi-Head Attention):} 
The first layer takes the 128-dimensional node features and applies multi-head attention to capture diverse interaction patterns.
\begin{itemize}[leftmargin=1.2em, itemsep=0.5pt, topsep=2pt]
    \item \textit{Input Channels:} $128$
    \item \textit{Heads:} $4$
    \item \textit{Hidden Channels per Head:} $8$
    \item \textit{Output Dimension:} The outputs of the $K$ heads are concatenated, resulting in a feature vector of size $4 \times 8 = 32$.
    \item \textit{Activation:} $\text{ReLU}(\cdot)$.
\end{itemize}

\paragraph{Layer 2 (Opinion Regression):} 
The second layer aggregates the hidden representations to regress the final opinion value.
\begin{itemize}[leftmargin=1.2em, itemsep=0.5pt, topsep=2pt]
    \item \textit{Input Channels:} $32$
    \item \textit{Heads:} $1$
    \item \textit{Output Channels:} $1$ (Scalar opinion value)
    \item \textit{Activation:} $\text{Tanh}(\cdot)$, to constrain the predicted opinion $\mathbf{o}^{t+1}$ within the valid opinion range of $[-1, 1]$.
\end{itemize}
Table~\ref{tab:gat_params} summarizes the above shapes and parameters of the network modules.

\begin{table}[h]
    \centering
    \small
    \begin{tabular}{l|c c}
    \toprule
    \textbf{Module} & \textbf{Parameter} & \textbf{Value/Shape} \\
    \midrule
    \multirow{2}{*}{Dynamic MLP} & Input Dim & 9 \\
                                 & Output Dim & 64 \\
    \midrule
    \multirow{2}{*}{Static MLP} & Input Dim & 768 \\
                                & Output Dim & 64 \\
    \midrule
    \multirow{4}{*}{GAT Layer 1} & Input Channels & 128 \\
                                 & Attention Heads & 4 \\
                                 & Hidden Units & 8 \\
                                 & Edge Dim & 1 \\
    \midrule
    \multirow{3}{*}{GAT Layer 2} & Input Channels & 32 \\
                                 & Attention Heads & 1 \\
                                 & Output Dim & 1 \\
    \midrule
    Output Activation & Function & Tanh \\
    \bottomrule
    \end{tabular}
    \caption{Hyperparameter settings and shapes for the GMP neural architecture.}
    \label{tab:gat_params}
\end{table}

\subsubsection{Training Design}
\label{appendix: GMP_training}
The GMP model is designed to recursively update the opinions of ordinary agents in parallel. By leveraging the forward pass of a Graph Attention Network (GAT), it captures fine-grained interaction dynamics. Specifically, the model requires only the initial opinion values of agents; the module then recursively constructs dynamic feature inputs based on historical states, along with the static agent profile features, to regress the future opinion values.

However, training this supervised model on real-world data presents significant challenges, primarily due to the irregularity of social media datasets. The three main obstacles are:
\begin{itemize}
    \item \textbf{Inconsistent Observation Space:} The number of participating users varies significantly across different event datasets, preventing a fixed-size input structure.
    \item \textbf{Discontinuous Opinion Trajectories:} Users typically post sporadically. Consequently, quantifiable opinion scores are often sparse and discontinuous over time, complicating the training of a recursive time-series model.
    \item \textbf{Multi-scale Alignment:} The objective function must balance the need to fit micro-level individual stances while simultaneously capturing the macro global opinion trend.
\end{itemize}

To address these challenges, we propose a comprehensive training scheme comprising three key components:

\paragraph{\textbf{a) User Embedding and Clustering}}
To resolve the issue of inconsistent observation spaces, we normalize the variable number of real users into a fixed set of "virtual agents." We crawl user profile descriptions across events, generate semantic embeddings via text encoding, and cluster these vectors into 1000 distinct classes. Each class represents a virtual agent. This approach ensures a consistent input dimension across diverse events and preserves privacy by preventing overfitting to specific real-world individuals.

\paragraph{\textbf{b) Trajectory Interpolation}}
To handle temporal discontinuity, we employ a hybrid interpolation strategy to construct continuous opinion trajectories from sparse data. For time steps lacking explicit stance labels, opinion values are imputed using a weighted combination: 50\% derived from linear interpolation of the specific agent's history, and 50\% sampled from a normal distribution based on the global opinion variance at that time step. This ensures the training data maintains temporal continuity and statistical consistency.

\paragraph{\textbf{c) Optimization Objective}}
To ensure the model captures both macro-level trends and micro-level diversity, the loss function is designed as a weighted sum of individual and global errors.

On one hand, the \textit{individual error} ($L_{\text{local}}$) minimizes the Mean Squared Error (MSE) between the predicted stance $\hat{\mathbf{o}}^{t}$ and the ground truth $\mathbf{o}^{t}$ for all virtual agents,
\begin{equation}
    L_{\text{local}} = \frac{1}{t_{max}}\sum_t \frac{1}{n} \left\| \hat{\mathbf{o}}^{t} - \mathbf{o}^{t} \right\|_2^2,
    \label{eq:individual_loss}
\end{equation}
where $t_{max}$ represents the training window, $n$ is the number of virtual agents, and $\left\|\cdot \right\|_2$ denotes the $L_2$ norm.

On the other hand, the \textit{global error} ($L_{\text{global}}$) ensures the predicted average opinion aligns with the macro ground truth,
\begin{equation}
    L_{\text{global}} = \frac{1}{t_{\max}} \sum_{t} \left( \frac{1}{n} \sum_{i=1}^{n} \hat{o}_i^{t} - \frac{1}{n} \sum_{i=1}^{n} o_i^{t} \right)^2.
    \label{eq:global_loss}
\end{equation}

Based on the above design, the total loss function is formulated as
\begin{equation}
    \mathcal{L} = \alpha \cdot L_{\text{local}} + \beta \cdot L_{\text{global}}.
    \label{eq:total_loss}
\end{equation}
To encourage the learning of distinct local features and prevent the model from collapsing individual predictions into the global mean, we assign weights $\alpha=0.9$ and $\beta=0.1$. 

It is important to note that, the objective of GMP is to learn relational influence weights among user nodes with heterogeneous dynamic features (opinion statistics) and static features (profiles) under varying network topologies, rather than to fit ground-truth individual opinion values. Specifically, GMP is trained on real user data from the first ten steps (1,000 user nodes) of the datasets, including user profiles and opinion states. It is subsequently evaluated in a simulation across the entire 30-step duration with 10,000 generated agents different from training data. These agents feature newly generated profiles, network structures, and opinion spaces where ordinary agents are influenced by dynamically generated opinions from core agents. This setup is analogous to training on a training set and performing inference on a separate test set, which inherently avoids data leakage.

\subsection{Empirical Assessment of EDG}
\label{appendix: EDG}

To provide direct structural evidence, we analyze the centrality of EDG-selected core agents within a 10,000-agent simulated network exhibiting a long-tailed degree distribution, where in-degree represents the number of followers (global statistics: min = 10, max = 244, mean = 19.43, p80 = 22.0).

As shown in Table~\ref{tab:EDG_Centrality}, an average of 94.1\% of the core agents selected per round fall within the top 20\% in-degree tier ($\ge$ p80). This aligns with the Pareto Principle (often referred to as the 80/20 rule), where a small fraction of agents exert the majority of influence. These results demonstrate that EDG preferentially selects influential nodes across all simulation rounds, which provides strong evidence supporting our claim that EDG identifies emergent opinion leaders.

\begin{table}[h]
    \centering
    \scriptsize
    \renewcommand{\arraystretch}{1.2} 
    \resizebox{\columnwidth}{!}{
    \begin{tabular}{c|c|c|c|c|c}
    \hline
    Round & $\ge$ p80 & p60-80 & p40-60 & <p40 & in Top 20\%\\
    \hline
    1 & 100 & 0 & 0 & 0 & 100\% \\
    5 & 97 & 3& 0 & 0 & 97\%\\
    10 & 88 & 11 & 1 & 0 & 88\%\\
    15 & 98 & 2 & 0 & 0 & 98\%\\
    20 & 83 & 15 & 2 & 0 & 83\%\\
    25 & 99 & 1 & 0 & 0 & 99\%\\
    30 & 96 & 4 & 0 & 0 & 96\%\\
    \hline
    Mean & - & - & - & - & \textbf{94.1\%} \\
    \hline           
    \end{tabular}
    }
    \caption{Centrality of EDG-selected agents.}
    \label{tab:EDG_Centrality}
\end{table}

\subsection{Heterogeneity Design}
\label{appendix: heterogeneity}
The heterogeneity in GASim is modeled at both the generative (LLM-driven reasoning of core agents) and parametric (GMP-based opinion updating of ordinary agents) levels, rather than assuming agent homogeneity. Each agent is initialized with a distinct persona profile sampled from real-world distributions (e.g., name, gender, nationality, personality, interests), which shapes susceptibility, stubbornness, and activity patterns.

When agents are dynamically selected as core agents by EDG, they generate responses via persona-conditioned LLM reasoning. The prompt explicitly enforces bias-consistent and personality-grounded behavior which requires agents to:
\begin{itemize}[itemsep=0.5pt, topsep=2pt]
    \item base decisions solely on provided context, including persona, trigger news, personal memory, and twitter page
    \item focus selectively on potential contradictions aligned with their concerns (see Appendix~\ref{appendix:bias_prompt})
    \item maintain personal biases unless sufficiently convinced
\end{itemize}

When acting as ordinary agents, heterogeneity is explicitly modeled by GMP. Beyond dynamic opinion states, GMP incorporates static persona embeddings encoded by BERT, enabling differentiated opinion update dynamics across agents with diverse traits.

\subsection{Statistical Report}

We conducted sensitivity analyses on key hyperparameters on the Politics dataset.
Specifically, we evaluate the hyperparameter choices of Top-K core agents (Top-K), entropy windowing of EDG (Window), memory graph kNN (KNN), and GMP architecture depth (Depth). By default, Top-K is set to be 100; Window is set to be 1; KNN is set to be 10; Depth is set to be 2. As shown in Table~\ref{tab:parameter}, performance remains stable under moderate variations, indicating that GASim is not overly sensitive to hyperparameter choices.

\begin{table}[h]
    \centering
    \small
    \begin{tabular}{lcccccc} 
        \toprule 
         & \textbf{Value} & \textbf{$\Delta$Bias} & \textbf{$\Delta$Div} & \textbf{Corr.} & \textbf{F.} \\
        \midrule 
        Default & -- & 0.0700 & 0.0074 & 0.4261 & 0.1349 \\
        \midrule
        Top-K & 50 & 0.0980 & 0.0096 & 0.3972 & 0.1485 \\
                          & 150 & 0.0808 & 0.0072 & 0.5376 & 0.1112 \\
        \midrule
        Window   & 2 & 0.0828 & 0.0086 & 0.3119 & 0.1784 \\
                          & 3 & 0.0797 & 0.0091 & 0.2430 & 0.2284 \\
        \midrule
        KNN  & 5 & 0.0746 & 0.0074 & 0.4011 & 0.1737 \\
                          & 15 & 0.0848 & 0.0096 & 0.2238 & 0.2107 \\
        \midrule
        Depth         & 1 & 0.0826 & 0.0098 & 0.2768 & 0.2130 \\
                          & 3 & 0.0775 & 0.0089 & 0.3122 & 0.1630 \\
        \bottomrule 
    \end{tabular}
    \caption{Results of sensitivity evaluation.} 
    \label{tab:parameter}
\end{table}

\subsection{Data Collection and Preprocessing}
\label{appendix: dataset}
We performed keyword-based retrieval to crawl event-related tweets from social media. For the opinion value annotation, we utilized the \texttt{gpt-4o-mini} model to assign a stance score to each tweet within the interval of $[-1, +1]$. In this scoring system, a value of $-1$ represents an \textit{extremely negative} stance toward the event, while $+1$ represents an \textit{extremely positive} stance. The LLM scorers used for the Politics\footnote{\url{https://en.wikipedia.org/wiki/Russian_interference_in_the_2016_United_States_elections}}, Business\footnote{\url{https://en.wikipedia.org/wiki/Xinjiang_cotton_industry}}, and Education\footnote{\url{https://finance.yahoo.com/news/student-wowed-china-alibaba-math-070107525.html}} simulation events in our dataset are as follows.

\begin{tcolorbox}[
    breakable,
    colback=gray!10, 
    colframe=gray!50, 
    arc=5pt,         
    boxrule=0.5pt,   
    left=10pt, right=10pt, top=10pt, bottom=10pt, 
    fontupper=\scriptsize\ttfamily
]
Based on the comment, output the attitude score in the discussions of whether Russia intervenes in the 2016 US election. \\
\\
-1 indicates completely disbelief in the target, thinking the whole thing as a political conspiracy from the other party and fully defending for Donald Trump. \\
1 indicates completely belief in the target and strongly condemning Trump and his teams. \\
\\
Only output a score (float number) in the range of [-1, 1].
\end{tcolorbox}

\begin{tcolorbox}[
    breakable,
    colback=gray!10, 
    colframe=gray!50, 
    arc=5pt,         
    boxrule=0.5pt,   
    left=10pt, right=10pt, top=10pt, bottom=10pt, 
    fontupper=\scriptsize\ttfamily
]
Based on the comment, output the attitude score in the discussions of whether China Xinjiang cotton is produced through forced labor. \\
\\
-1 indicates complete disbelief in the target, strongly defending for China, and thinking the whole thing as a political conspiracy from western countries.\\
1 indicates complete belief in the target and strongly condemning the human rights violation in Xinjiang.\\
\\
Only output a score (float number) in the range of [-1, 1].
\end{tcolorbox}

\begin{tcolorbox}[
    breakable,
    colback=gray!10, 
    colframe=gray!50, 
    arc=5pt,         
    boxrule=0.5pt,   
    left=10pt, right=10pt, top=10pt, bottom=10pt, 
    fontupper=\scriptsize\ttfamily
]
Based on the comment, output the attitude score in the discussions of whether "Jiang Ping", a girl who is the first technical secondary school student from China that won the 12th place in the Alibaba Global Math Competition, had potentials cheating and the whole thing was a over-hyped publicity by made by the girl, her teacher and Alibaba. \\
\\
-1 indicates completely disbelieving Jiang Ping, thinking the whole thing as a media hype, or expressing extreme negative emotions (such as strong dissatisfaction, criticism, attacks, or sarcasm). \\
\\
1 indicates expressing strong defend, protection and support for Jiang Ping in terms of her gender and potential hard work.\\
Only output a score (float number) in the range of [-1, 1].
\end{tcolorbox}

Following the crawling and scoring phases, the tweet data underwent comprehensive cleaning and organization. Each processed data record contains the following attributes,
\begin{itemize}[itemsep=0.5pt, topsep=2pt]
    \item \textbf{User ID}: The unique identifier of the user account.
    \item \textbf{User Description}: The profile biography or description provided by the user.
    \item \textbf{Follower Count}: The total number of followers the user has.
    \item \textbf{Following Count}: The number of accounts the user is following.
    \item \textbf{Tweet Content}: The original text published by the user.
    \item \textbf{Posting Time}: The specific timestamp when the tweet was published.
    \item \textbf{Opinion Value}: The polarity value in $[-1, +1]$ derived from the LLM-based annotation.
\end{itemize}

\subsection{Metrics for Trend Alignment}
\label{appendix: macro_metrics}

To quantitatively evaluate the alignment between simulated and real-world public opinion, we define the simulated trend curve $S$ as the sequence of average agent stances at each time step $t \in \{1, \dots, t_{\max}\}$. Let $n$ be the number of agents and $o^t_i$ be the stance of agent $i$ at time $t$; the simulated average is $o^t_{sim} = \frac{1}{n} \sum_{i=1}^{n} o^t_i$. We compare $S=\{o^1_{sim}, \dots, o^{t_{\max}}_{sim}\}$ against the ground-truth curve $G = \{o^1_{global}, \dots, o^{t_{\max}}_{global}\}$ using the following four metrics.

\paragraph{$\boldsymbol{\Delta{\textbf{Bias}}}$ (Mean Absolute Bias).} 
This metric assesses the macro-level magnitude of error by measuring the average deviation between the simulated stance and the ground-truth labels across the entire duration. A lower value indicates higher simulation accuracy. It is defined as
\begin{equation}
\Delta\text{Bias} = \frac{1}{t_{\max}} \sum_{t=1}^{t_{\max}} |o^t_{global} - o^t_{sim}|.
\end{equation}

\paragraph{$\boldsymbol{\Delta{\textbf{Div}}}$ (Variance of Absolute Bias).}
To capture the temporal stability of the simulation, we compute the variance of the absolute error. This metric reflects whether the simulation error remains consistent or fluctuates significantly over time. It is defined as
\begin{equation}
\Delta\text{Div} = \frac{1}{t_{\max}} \sum_{t=1}^{t_{\max}} (|o^t_{global} - o^t_{sim}| - \Delta\text{Bias})^2.
\end{equation}

\paragraph{\textbf{Corr.} (Pearson Correlation Coefficient).}
The $\text{Corr.}$ metric measures the linear alignment and trend consistency between the two sequences. It evaluates how well the simulated fluctuations track the real-world rises and falls in public opinion. It is defined as
\begin{equation}
\text{Corr.} 
=
\frac{\operatorname{Cov}\!\left(o_{\text{global}},\, o_{\text{sim}}\right)}
{\sqrt{\operatorname{Var}\!\left(o_{\text{global}}\right)\operatorname{Var}\!\left(o_{\text{sim}}\right)}}.
\end{equation}

\paragraph{\textbf{F.} (Fréchet Distance).}
As a geometric metric, the Fréchet distance \textbf{F.} measures the global similarity between curves $S$ and $G$ by considering the location and ordering of points. It calculates the minimized maximum distance required to traverse both curves simultaneously. It is defined as
\begin{equation}
\textbf{F.} = \inf_{\alpha, \beta} \max_{t \in [0, 1]} \|G(\alpha(t)) - S(\beta(t))\|.
\end{equation}
where $\alpha(t)$ and $\beta(t)$ are continuous, non-decreasing reparameterization functions that map the normalized time interval $[0, 1]$ onto the curves' domains. Unlike simpler metrics, \textbf{F.} accounts for both temporal alignment and geometric shape, providing a more stringent assessment of structural similarity than Dynamic Time Warping (DTW).


\subsection{Baseline Methods}
\label{appendix: baseline}

\subsubsection{Trend Alignment Evaluation}
\label{appendix: baseline_alignment}
For the social simulation experiments, we consider two categories of baseline methods: traditional agent-based models (ABMs) and LLM-based social simulation frameworks.

\paragraph{Agent-Based Models.}
{HK}~\citep{HK}, {RA}~\citep{RA}, and {Lorenz}~\citep{Lorenz} are classic mathematical ABMs that describe attitude evolution through local interactions among neighboring agents. These models share a common computational paradigm consisting of neighbor selection, message exchange, and attitude update functions. Detailed formulations and comparative analyses are provided in~\citep{hisim}.

\paragraph{LLM-Based Frameworks.}
{SOD}~\citep{SOD} is an LLM-driven framework for modeling belief formation and confirmation bias. It shows that LLM agents naturally converge toward factual consensus, while carefully designed prompts can induce opinion fragmentation, making it a strong baseline for opinion dynamics.
{HiSim}~\citep{hisim} is the first hybrid simulation framework that integrates LLM-driven agents with traditional ABMs to significantly scale social simulations, serving as another competitive baseline for trend alignment evaluation.

\subsubsection{Memory Architecture Evaluation}
\label{appendix: baseline_memory}
To evaluate the effectiveness of our memory design on the LoCoMo retrieval benchmark, we compare against representative memory architectures that have previously been evaluated on this dataset.

\paragraph{A-Mem.\cite{A-mem}}
A-Mem introduces an agentic memory system for LLM agents that represents experiences as interconnected notes. Each note captures interactions enriched with structured attributes such as keywords, contextual descriptions, and tags generated by the LLM. New memories retrieve relevant notes using semantic embeddings and establish links through LLM-based similarity reasoning. Existing notes are dynamically updated as new information is integrated, allowing the memory structure to evolve and support increasingly rich contextual associations. Memory retrieval is performed via semantic similarity to provide relevant historical context during agent interactions.

\paragraph{Zep.\cite{zep}}
Zep is an memory-layer service powered by Graphiti, a dynamic and temporally aware knowledge graph engine. It synthesizes unstructured conversational data and structured business data into a non-lossy knowledge graph. Unlike static representations, Zep explicitly maintains the temporal validity of facts and relationships, enabling robust modeling of evolving environments.

\paragraph{LangMem.} 
LangMem \footnote{\url{https://langchain-ai.github.io/langmem/}} is a open-source memory architectures for long-term agent memory and behavioral adaptation. It automates knowledge extraction and consolidation from conversations and employs a background manager to continuously update agent state. Integrated with LangGraph, LangMem enables agents to maintain consistent and personalized behavior through functional memory primitives and iterative prompt refinement.

\paragraph{Mem0 and Mem0g.\cite{mem0}}
Mem0 manages agent memory by extracting salient information and using an LLM to perform \textsc{Add}, \textsc{Update}, \textsc{Delete}, or \textsc{NoOp} operations based on semantic comparisons with existing vector-embedded records. Mem0g extends this design to a graph-based architecture with a two-stage pipeline that extracts entities and relational triplets. It preserves temporal context through semantic node merging and an LLM-based conflict resolution mechanism that marks outdated relationships as invalid rather than deleting them, supporting complex relational reasoning over time.

\subsection{Supplementary Prompts}

\subsubsection{News Event Injection}
We perform different \textit{news event injections} at predefined time steps to simulate external information shocks across topic-specific events in our dataset, as detailed below.

The following prompt contains the news triggers for the Politics dataset.

\begin{tcolorbox}[
  breakable, 
  colback=gray!10,
  colframe=gray!50,
  arc=5pt,
  boxrule=0.5pt,
  left=10pt,right=10pt,top=10pt,bottom=10pt,
  fontupper=\scriptsize\ttfamily
]
\noindent
tr\_trigger\_news1: \&tr\_trigger\_news1\\
Former President's National Security Advisor Flynn resigned three
weeks into office for discussing Russia sanctions with the Russian
ambassador and concealing it from the Vice President. FBI Director
Comey was fired by President Trump, after sources revealed he had
recently asked the Justice Department for more funds to investigate
Russia's involvement in the 2016 U.S. election.\\
\\
tr\_trigger\_news2: \&tr\_trigger\_news2\\
In the context of a presidential election, the ``Russiagate''
controversy appears to have become highly politicized, which may
gradually turn into a campaign issue repeatedly used by both
parties to mobilize supporters and attack each other. \\
\\
tr\_trigger\_news3: \&tr\_trigger\_news3\\
Jared Kushner, director of the White House Innovation Office,
son-in-law and senior adviser to President Trump, issued a
statement in Washington to explain his four meetings with Russian
officials during Trump's campaign and transition period before
taking office. He said he had never had ``inappropriate contacts''
with foreign countries.\\
\\
tr\_trigger\_news4: \&tr\_trigger\_news4\\
U.S. Special Prosecutor Mueller announced that Trump’s former
campaign manager Manafort and three others were prosecuted for
improper contacts with Russia and suspected crimes, including
money laundering, conspiracy, perjury, false reporting, and
concealing bank information. The accused are George
Papadopoulos, Paul Manafort, and Rick Gates.\\
\\
tr\_trigger\_news5: \&tr\_trigger\_news5\\
In Washington, U.S., Michael Flynn, former national security
adviser to U.S. President Trump, appeared in court and pleaded
guilty to perjury to the FBI in the Russia investigation.
\end{tcolorbox}

Similarly, the following prompt contains the news triggers for the Business dataset.

\begin{tcolorbox}[
  breakable, 
  colback=gray!10,
  colframe=gray!50,
  arc=5pt,
  boxrule=0.5pt,
  left=10pt,right=10pt,top=10pt,bottom=10pt,
  fontupper=\scriptsize\ttfamily
]
\noindent
xj\_trigger\_news1: \&xj\_trigger\_news1\\
1 - Reports reveal that Xinjiang cotton is produced through forced labor. ABC's Four Corners investigative program said Uighur Muslims were arrested and forced to work in textile factories in Xinjiang. \#XinjiangCotton\\
2 - (Breaking News) The well-known clothing brand H\&M announced that it has terminated its relationship with a yarn supplier in mainland China (Huafu Fashion) because the factory's products were suspected of being produced using ``forced labor'' of ethnic minorities in Xinjiang. Soon after Chinese state media and mainland Chinese netizens jointly boycott H\&M.\\
\\
xj\_trigger\_news2: \&xj\_trigger\_news2\\
Hua Chunying, spokesperson for China's Ministry of Foreign Affairs, responded at a press conference, stating that Xinjiang's cotton is among the best in the world and that the ``forced labor'' accusation is a malicious lie fabricated by anti-China forces to damage China's image and Xinjiang's stability. She emphasized that China is open and transparent, but the will of the Chinese people cannot be deceived, noting that over 40\% of Xinjiang's cotton fields are mechanized, countering Western criticism.\\
\\
xj\_trigger\_news3: \&xj\_trigger\_news3\\
H\&M faced widespread backlash in China for its stance on Xinjiang cotton, leading to the termination of partnerships, store closures, and a significant decline in sales, with China dropping out of the brand's top markets by mid-2021.\\
\\
xj\_trigger\_news4: \&xj\_trigger\_news4\\
On March 31, H\&M affirmed its long-term commitment to the Chinese mainland market and its goal to be a responsible buyer globally, emphasizing collaboration with stakeholders to develop the fashion industry. However, the response did not address Xinjiang and was not well received by Chinese netizens.\\
\\
xj\_trigger\_news5: \&xj\_trigger\_news5\\
On April 15, 2021, CGTN reported that the BCI official website removed the statement that ``human rights violations and forced labor exist in Xinjiang''. However, when The Economist's China editor consulted BCI, BCI said its policy remained unchanged and that it would only reissue the notice in response to attacks on its website.
\end{tcolorbox}

Moreover, the following prompt contains the news triggers for the Education dataset.

\begin{tcolorbox}[
  breakable, 
  colback=gray!10,
  colframe=gray!50,
  arc=5pt,
  boxrule=0.5pt,
  left=10pt,right=10pt,top=10pt,bottom=10pt,
  fontupper=\scriptsize\ttfamily
]
\noindent
jp\_trigger\_news1: \&jp\_trigger\_news1\\
1 - Don't look down on the poor youngsters! A technical secondary school girl Jiang Ping unexpectedly won the 12th place in the world in the mathematics competition.\\
2 - There is an online rumor that Harvard University, the University of Hong Kong and many other well-known universities were going to admit the vocational school student Jiang Ping through an exceptional admission process, but the official has refuted the rumor.\\
\\
jp\_trigger\_news2: \&jp\_trigger\_news2\\
1 - How disrespectful to Jiang Ping! The Ali finals just ended, 39 finalists issued a joint letter questioning Jiang Ping.\\
2 - Shocking Viral News - famous ``Anti-counterfeiting fighter'' Fang Zhouzi questioned Jiang Ping, as her math score in the high school entrance exam was only 83 points (full score is 150 points), which was out of place when compared to the rankings in global competitions. Besides, in the promotional video released by Alibaba Damo Academy, her blackboard writing seemed to have many low-level mistakes that should not be made by an excellent math learner. Moreover, the competition is an open book test, so cheating is not difficult. \#Jiang Ping \#Fang Zhouzi\\
\\
jp\_trigger\_news3: \&jp\_trigger\_news3\\
1 - Jiang Ping dropped out of school? Lianshui Middle School internal news leaked -> Has the genius girl ``fallen''?\\
2 - Official notice- the online rumor that Jiang Ping failed the monthly math test in school is true.\\
\\
jp\_trigger\_news4: \&jp\_trigger\_news4\\
Official notice- According to the investigation, Jiang Ping was provided help by her teacher Mr. Wang during the preliminaries, which violated the preliminaries rule of ``prohibiting discussion with others''.
\end{tcolorbox}


\subsubsection{Agent Cognitive Bias}
\label{appendix:bias_prompt}
In this appendix, we detail the configuration of cognitive biases for the LLM-driven core agents within our simulation framework. To mimic the polarized nature of real-world social media discussions, agents are not treated as neutral information processors. Instead, they are assigned specific cognitive frameworks—referred to here as Potential Contradictions—which dictate their initial stances and their resistance to opposing information.

\begin{tcolorbox}[
  breakable, 
  colback=gray!10,
  colframe=gray!50,
  arc=5pt,
  boxrule=0.5pt,
  left=10pt,right=10pt,top=10pt,bottom=10pt,
  fontupper=\scriptsize\ttfamily
]
\textbf{Agent Behavioral Instruction:} \
\textit{``You should speak like a real person with personal biases regarding the news. Only when the proof and news are convincing will you alter your opinion. When you hear opinions that are opposed to your own, you might tend to choose one of the 'potential contradictions' that concerns you and formulate biased views.''}
\end{tcolorbox}

For the Politics dataset, the bias prompt models the partisan divide characteristic of U.S. political discourse.
\begin{tcolorbox}[
  breakable, 
  colback=gray!10,
  colframe=gray!50,
  arc=5pt,
  boxrule=0.5pt,
  left=10pt,right=10pt,top=10pt,bottom=10pt,
  fontupper=\scriptsize\ttfamily
]
(Democrats and Supporters): They believe that Trump's interactions and subsequent cover-up regarding Russia are sufficient in themselves to pose serious political and legal problems, and they believe that Trump's team may have improperly colluded with Russia during the election.\\
\\
(Trump and Supporters): They believe that the ``Russiagate'' is merely a political conspiracy aimed at undermining the legitimacy of the Trump administration and the presidency. They argue that there is insufficient evidence to prove any illegal collusion between Trump and Russia.
\end{tcolorbox}

For the Education dataset, the bias prompt captures the intersection of academic elitism, gender prejudice, and social class.

\begin{tcolorbox}[
  breakable, 
  colback=gray!10,
  colframe=gray!50,
  arc=5pt,
  boxrule=0.5pt,
  left=10pt,right=10pt,top=10pt,bottom=10pt,
  fontupper=\scriptsize\ttfamily
]
(Education Conflict): When Jiang Ping is being questioned by the authority, do you tend to defend Jiang Ping and express dissatisfaction with exam-oriented education, academic qualifications, and class stratification considering her status as a technical secondary school student, poor family background, and passion for learning?\\
\\
(Gender Conflict): Do you tend to defend Jiang Ping by believing that the suspicion toward her is rooted in prejudice against women, and that there would not be so many doubts if the contestant were a man?\\
\\
(Class Conflict): Do you tend to think it is impossible for Jiang Ping to achieve such accomplishment considering the possible evidence of cheating and her educational background?\\
\\
(Media Moral Conflict): Do you tend to think that the entire event is essentially a staged media hype designed for viral engagement?
\end{tcolorbox}

For the Business dataset, the bias prompt represents the clash between international human rights narratives and domestic nationalist sentiment.
\begin{tcolorbox}[
  breakable, 
  colback=gray!10,
  colframe=gray!50,
  arc=5pt,
  boxrule=0.5pt,
  left=10pt,right=10pt,top=10pt,bottom=10pt,
  fontupper=\scriptsize\ttfamily
]
(Western Media and Business Circles): They accuse Chinese authorities of imposing ``forced labor'' on the Uyghur people in Xinjiang and launch trade sanctions and boycotts against products sourced from the region.\\
\\
(Chinese Official Media and Public): They accuse the West of malicious smearing and disinformation, mobilizing widespread support for Xinjiang cotton while initiating boycotts against companies that abandoned the regional supply chain.
\end{tcolorbox}

\subsubsection{Memory Judge Prompt}
\label{appendix: memory_judge}

For the memory architecture evaluation based on LoCoMo dataset, we adopt the memory judge prompt as follows,
\begin{tcolorbox}[
  breakable, 
  colback=gray!10,
  colframe=gray!50,
  arc=5pt,
  boxrule=0.5pt,
  left=10pt,right=10pt,top=10pt,bottom=10pt,
  fontupper=\scriptsize\ttfamily
]
Your task is to label an answer to a question as ’CORRECT’ or ’WRONG’. You will be given the following data:
    (1) a question (posed by one user to another user), 
    (2) a ’gold’ (ground truth) answer, 
    (3) a generated answer
which you will score as CORRECT/WRONG. \\
\\
The point of the question is to ask about something one user should know about the other user based on their prior conversations.
The gold answer will usually be a concise and short answer that includes the referenced topic, for example: \\
Question: Do you remember what I got the last time I went to Hawaii? \\
Gold answer: A shell necklace \\
The generated answer might be much longer, but you should be generous with your grading - as long as it touches on the same topic as the gold answer, it should be counted as CORRECT. \\
\\
For time related questions, the gold answer will be a specific date, month, year, etc. The generated answer might be much longer or use relative time references (like "last Tuesday" or "next month"), but you should be generous with your grading - as long as it refers to the same date or time period as the gold answer, it should be counted as CORRECT. Even if the format differs (e.g., "May 7th" vs "7 May"), consider it CORRECT if it's the same date. \\
\\
Now it's time for the real question: \\
Question: \{question\} \\
Gold answer: \{gold\_answer\} \\
Generated answer: \{generated\_answer\} \\
\\
First, provide a short (one sentence) explanation of your reasoning, then finish with CORRECT or WRONG. 
Do NOT include both CORRECT and WRONG in your response, or it will break the evaluation script. \\
\\
Just return the label CORRECT or WRONG in a json format with the key as "label".
\end{tcolorbox}

\end{document}